\documentclass{article}
\usepackage{graphicx,geometry,amsmath,amssymb,caption,subcaption,enumerate,url,xcolor,lmodern,pdftexcmds,authblk}

\DeclareMathAlphabet{\mathsfbi}{OT1}{\sfdefault}{bx}{sl}
\DeclareMathVersion{sfletters}
\SetSymbolFont{letters}{sfletters}{OML}{ntxsfmi}{b}{it}

\makeatletter
\newcommand{\tensorsbilow}[1]{%
  \text{\mathversion{sfletters}$\m@th#1$}%
}

\DeclareRobustCommand{\tensor}[1]{%
  \begingroup
  \ifcat\noexpand #1\relax
    \edef\greek@test{\detokenize{#1}}%
    \edef\greek@test{\expandafter\@cdr\greek@test\@nil}%
    \edef\greek@test{\expandafter\@car\greek@test\@nil}%
    \edef\x{\the\lccode\expandafter`\greek@test}%
    \edef\y{\number\expandafter`\greek@test}%
    \ifnum\x=\y\relax
      \tensorsbilow{#1}%
    \else
      \mathsfbi{#1}%
    \fi
  \else
    \mathsfbi{#1}%
  \fi
  \endgroup
}
\makeatother

\newcommand{\bx}{\tensor{x}}
\newcommand{\bD}{\tensor{D}}
\newcommand{\bX}{\tensor{X}}
\newcommand{\RR}{\mathbb{R}}
\newcommand{\EE}{\mathbb{E}}
\newcommand{\mD}{\mathcal{D}}
\newcommand{\bC}{\tensor{C}}
\newcommand{\by}{\tensor{y}}
\newcommand{\bz}{\tensor{z}}
\newcommand{\bbeta}{\tensor{\beta}}
\newcommand{\bK}{\tensor{K}}
\newcommand{\bk}{\tensor{k}}

\newcommand{\bal}{\tensor{\alpha}}
\newcommand{\bGam}{\tensor{\Gamma}}
\newcommand{\bl}{\tensor{l}}
\newcommand{\bL}{\tensor{L}}
\newcommand{\bLam}{\tensor{\Lambda}}
\newcommand{\bmu}{\tensor{\mu}}
\newcommand{\bm}{\tensor{m}}
\newcommand{\mO}{\mathcal{O}}
\newcommand{\bw}{\tensor{w}}
\newcommand{\bA}{\tensor{A}}
\newcommand{\bSigma}{\tensor{\Sigma}}
\newcommand{\bPhi}{\tensor{\Phi}}
\newcommand{\bphi}{\tensor{\phi}}
\newcommand{\dx}{\mathrm{d}x}
\newcommand*{\keywords}[1]{\textbf{\textit{Keywords---}} #1}

\newcommand{\argmin}{\operatornamewithlimits{argmin}}
\newtheorem{thm}{Theorem}

\title{Analytical results for uncertainty propagation through trained machine learning regression models}
\author{Andrew Thompson\footnote{email: andrew.thompson@npl.co.uk\\ \indent © 2024. This manuscript version is made available under the CC-BY-NC-ND 4.0 license\\ \indent https://creativecommons.org/licenses/by-nc-nd/4.0/}}
\date{%
    National Physical Laboratory, Hampton Road, Teddington, TW11 0LW, UK\\[2ex]%
    \today}

\begin{document}

\maketitle

\begin{abstract}
Machine learning (ML) models are increasingly being used in metrology applications. However, for ML models to be credible in a metrology context they should be accompanied by principled uncertainty quantification. This paper addresses the challenge of uncertainty propagation through trained/fixed  ML  regression models. Analytical expressions for the mean and variance of the model output are obtained/presented for certain input data distributions and for a variety of ML models. Our results cover several popular ML models including linear regression, penalised linear regression, kernel ridge regression, Gaussian Processes (GPs), support vector machines (SVMs) and relevance vector machines (RVMs). We present numerical experiments in which we validate our methods and compare them
with a Monte Carlo approach from a computational efficiency point of view. We also illustrate
our methods in the context of a metrology application, namely modelling the state-of-health
of lithium-ion cells based upon Electrical Impedance Spectroscopy (EIS) data.
\end{abstract}

    

\section{Introduction}\label{sec:intro}

Machine learning (ML) measurement models are increasingly being used in a wide range of metrology applications, for example in thermometry~\cite{bilson2023machine}, battery state-of-health modelling~\cite{chan2022comparison}, nanoparticle characterisation~\cite{coquelin2019towards}, earth observation~\cite{lary2018machine} and oceanography~\cite{robinson2023impact}. The appeal of ML is its ability to learn measurement models from data, even in cases where physical models are either not well understood or inefficient to compute. It also enables the automation of time-consuming processes that previously needed to be performed manually.

\subsection{Uncertainty quantification for ML models}

For ML approaches to be credible in a metrology context, it is important that they are accompanied by principled uncertainty quantification. The importance of uncertainty quantification for ML and the need for research on this topic was highlighted in the recent Strategic Research Agenda developed by the European Metrology Network for Mathematics and Statistics (MATHMET)~\cite[Section 3]{mathmet2024SRA}.  This paper addresses the challenge of uncertainty propagation through trained/fixed ML  regression models. In other words, given an observation of the input variables accompanied by knowledge of their corresponding uncertainty, the aim is to characterise the uncertainty of the ML model output. We choose to focus on regression models because they occur particularly frequently in a metrology context.

A framework for evaluating uncertainties by means of measurement models was standardised for the metrology community in the influential ``Guide to the Expression of Uncertainty in Measurement'' (GUM) and related supplements~\cite{balazs2008,bipm2008,bipm2008supplement,bipm2011}. The GUM paradigm mainly addresses the propagation of uncertainty through fixed models. However, since an ML model is learned from data, the absence of full physical insight leads to uncertainty in the model itself. Indeed, two types of uncertainty are typically distinguished in the ML community: {\emph{data uncertainty} and \emph{model uncertainty}. Data uncertainty is inherent to the modelling task, and so is in this sense irreducible. Model uncertainty, on the other hand, is uncertainty concerning the model itself, for example the form of the model or the value of its parameters. The need to take account of model uncertainty is in fact explicitly acknowledged within the GUM documents; see~\cite[3.1.6]{bipm2008} for example. It follows that a comprehensive uncertainty evaluation for ML models requires model uncertainty to be taken into account, which in turn also requires a consideration of the uncertainty in the training data. 

In this paper, we restrict our focus to the propagation of uncertainty through fixed ML models. While not constituting a comprehensive uncertainty evaluation, this challenge is still of interest for the following reasons:
\begin{enumerate}[(i)]
    \item It is useful to be able to disentangle the contributions of different sources of uncertainty: in this case to isolate the prediction uncertainty due to input variable uncertainty.
    \item Viewing the ML model as fixed fits well with the GUM framework, which is well understood by the metrology community.
    \item Rather than considering the output of an ML model to be an estimate of an underlying measurand, an alternative approach is to consider the output of the fixed ML model itself to be the measurand. In this case, the propagation of input uncertainty through the fixed model represents a comprehensive uncertainty evaluation.
\end{enumerate}

\subsection{Analytical approaches for uncertainty quantification}

The GUM framework identifies two main types of method for propagating uncertainties through measurement models: analytical and sampling-based. In the analytical approach, a mathematical expression is derived either for the distribution of the output or some expression of uncertainty such as variance. In the sampling-based approach, the model is evaluated on Monte Carlo samples of the input data distribution, thereby obtaining an empirical distribution.

If it is possible to derive analytical expressions characterising the output uncertainty, it is usually preferable to use them for several reasons:
\begin{enumerate}[(i)]
    \item \textbf{Accuracy.} Analytical expressions often precisely characterise the output uncertainty. This is in contrast to sampling-based approaches which only approximate the true distribution of the model output with accuracy increasing with the number of samples, which results in information loss that may be furthermore difficult to quantify.
    \item \textbf{Transparency.} Analytical expressions often provide a greater degree of transparency by providing insight and understanding into the sources of propagated uncertainty.
    \item \textbf{Computational efficiency.} It is often more computationally efficient to evaluate an analytical expression than to perform Monte Carlo sampling. 
    \item \textbf{Reproducibility.} Analytical approaches lead to uncertainty evaluations which are more easily reproducible. A characterisation of the input data and its uncertainty along with an analytical expression for its propagation is all that is needed to reproduce the evaluation. In contrast, for a Monte Carlo sampling approach, reproducibility would require, for example, the storage of all quantities sampled within the Monte Carlo method.
\end{enumerate}

\subsection{Our contribution}

In this paper, we focus on analytical characterisations of output uncertainty for some popular ML regression models. More specifically, we present uncertainty propagation results for two families of ML models: linear models and kernel-based models. These model families cover several popular ML approaches, including linear regression, ridge regression, kernel regression, support vector machines and Gaussian Processes. In each case, we give analytical expressions for the mean and variance of predictions from these models given certain kinds of information about input variable uncertainty. In some cases the results we present are novel, and we supplement these with known results in order to give as complete a presentation as possible. 

We note that, in contrast to sampling-based approaches, the results presented in this paper are generally restricted to the moments of the predictions and not other characterisations of uncertainty such as confidence/credible intervals or quantiles. Distributional assumptions on the output distributions would need to be made to obtain such expressions of uncertainty based on the results in this paper.

We demonstrate the practicality of our methods by illustrating their use within a metrology application, namely modelling the state-of-health of lithium-ion cells based upon Electrical Impedance Spectroscopy (EIS) data (see Section~\ref{batteries}). By comparing our methods with a Monte Carlo sampling approach in this context, we validate the correctness of our expressions.

We also carry out an investigation into the computational efficiency tradeoffs between the analytical and Monte Carlo sampling approaches. We discover that, for a given model, there exists an accuracy threshold below which a Monte Carlo sampling approach becomes more computationally expensive. We characterise the relationship between this threshold and certain parameters of the model, namely the number of input variables and the number of training samples. The analytical expressions, on the other hand, are exact.

\subsection{Paper structure}

The structure of the paper is as follows. We first review closely related work in Section~\ref{related}. Sections~\ref{linear_models} and~\ref{kernel_models} focus on two main families of models, namely linear models and kernel-based models respectively. In both sections, we first give an overview of the types of models within the family before presenting relevant analytical results for uncertainty propagation. Proofs for all the novel results in Section~\ref{kernel_models} can be found in Appendix~\ref{kernel_proofs}. We present our numerical experiments in Section~\ref{experiments}. We begin the section by giving background on state-of-health estimation for lithium-ion cells, before presenting various experiments validating the accuracy and computational efficiency of our analytical results through a comparison with a Monte Carlo sampling approach. We give our conclusions in Section~\ref{discussion}.

\subsection{Notation}

Before proceeding, we introduce some notation. Suppose we have a standard fixed regression model $y=f(\bx)$ which takes a vector of inputs $\bx\in\RR^m$ to a scalar output $y\in\RR$. Note that the output variable is real-valued since we are considering a regression model, and we also here exclude the possibility of categorical input variables. Then suppose that we wish to evaluate the model on some $\bX^{\ast}$, a random variable whose distribution is known; let us write $\bX^{\ast}\sim\mathcal{D}$. We then wish to characterize the distribution of the output of the model, namely the random variable $Y^{\ast}=f(\bX^{\ast})$. At times we will write $\bx^{\ast}$ and $y^{\ast}$ for (deterministic) instances of these random variables. 

We will also be interested at points in the training set from which the model was learned. We assume that a training set consisting of $n$ paired observations of inputs and outputs is available. We denote by $\bx_i\in\RR^m$ the observed vector of input variables for observation $i$ and we denote by $y_i$ the corresponding observed output variable, where $i=1,\ldots,n$. We write $\bC_{\bx}$ for the $n\times m$ \emph{data matrix} whose rows are the input data observations and we write $\by$ for the vector whose entries are the output data observations, that is
$$\bC_{\bx}:=\begin{bmatrix}\bx_1^{\top}\\ \vdots\\ \bx_n^{\top}\end{bmatrix},\;\;\;\;\by:=\begin{bmatrix}y_1\\ \vdots\\ y_n\end{bmatrix}.$$

In order to simplify the mathematics, we assume that the $\{\bx_i\}$ and the $\{y_i\}$ have been centered around their mean values $\bar{\bx}=\sum_{i=1}^n \bx_i$ and $\bar{y}=\sum_{i=1}^n y_i$ respectively.

\section{Related work}\label{related}

There has been some work in the metrology community in which input uncertainties have been propagated through fixed ML models. For example, in~\cite{venton2021robustness},  the robustness of convolutional neural networks for classifying cardiovascular disorders based on electrocardiogram (ECG) signals to physiological input noise was investigated. Analytical methods were not explored in these works however.

There is also a body of work in the metrology community in which input data uncertainties are taken into account when training linear models. These approaches include total least-squares~\cite{forbes2002generalised} and errors-in-variables~\cite{klauenberg2022gum}. The errors-in-variables approach was recently extended to neural networks in~\cite{martin2023aleatoric}.

The results concerning kernel-based models presented in Section~\ref{kernel_results} are all inspired by those in~\cite{candela2003propagation, quinonero2003prediction, wan2014analytical}, where results for combined propagated input data uncertainty and model uncertainty were obtained for Gaussian Processes  and  radial basis function (RBF) kernels. We extend these results in several directions.
\begin{enumerate}[(i)]
    \item We obtain results for fixed models through a simplification of their argument, essentially removing the model uncertainty terms in the expression (Theorem~\ref{general_kernel} in Section~\ref{kernel_results}). 
    \item We extend these results for fixed models to other models, namely kernel ridge regression and support vector machines (Theorem~\ref{general_kernel} in Section~\ref{kernel_results}). 
    \item We also obtain results for the case where the input random variables are independently distributed according to triangular  distributions (Theorem~\ref{kernel_tri} in Section~\ref{kernel_results})

    \end{enumerate}

\section{Linear models}\label{linear_models}

Some ML regression methods result in linear models, and in this case it is particular straightforward to analytically propagate input uncertainties. Methods that fall into this category are standard linear regression (ordinary least squares), penalised linear regressions such as ridge regression and sparse regression, and linear support vector machines. All of these methods return a point prediction of the form $y^{\ast}=\bbeta^{\top}\bx^{\ast}$, where $\bbeta$ is a vector of linear model weights. We briefly review each of the above-mentioned methods in Section~\ref{linear_review} and show that this is the case.

\subsection{Background on linear models}\label{linear_review}

Linear models are widely used in statistical modelling since they are easy to interpret and computationally cheap to compute. 

In standard linear regression, a linear model $y=\bbeta^{\top}\bx$ is learned by minimising a least-squares objective, that is
\begin{equation}\label{linear}
\bbeta:=\displaystyle\argmin_{\tilde{\bbeta}\in\RR^m}\|\by-\bC_{\bx}\tilde{\bbeta}\|_2^2,
\end{equation}
 for which a closed-form solution exists~\cite[Equation (3.6)]{hastie2009elements}. 

Note that necessary conditions for (\ref{linear}) to have a well-defined solution is that the number of observations $n$ is greater than or equal to the variable dimension $m$ -- a condition that is usually met in a data-rich ML context -- and that the matrix $\bC_{\bx}^{\top}\bC_{\bx}$ is full rank. 

In linear ridge regression, an $l_2$ regularisation term is added to the objective in (\ref{linear}). The regularisation term promotes smoothness in the solution and can help to prevent overfitting. It also allows for a well-defined solution even in the case of $n<m$. The optimisation problem
\begin{equation}\label{ridge}
\bbeta:=\displaystyle\argmin_{\tilde{\bbeta}\in\RR^m}\|\by-\bC_{\bx}\tilde{\bbeta}\|_2^2+\sigma^2\|\tilde{\bbeta}\|_2^2
\end{equation}
is solved, where $\sigma^2\geq 0$ is a regularisation parameter,  and this also has a closed-form solution~\cite[Equation (3.44)]{hastie2009elements}.

Other regularisers are also possible, for example an $l_1$-norm regulariser, often referred to as the LASSO~\cite{tibshirani1996regression}, which promotes sparsity in the weights $\bbeta$. In this case, no closed-form solution exists, but the unique solution of the convex optimisation problem can be obtained using efficient algorithms.

Support vector machines (SVMs) are a popular ML approach for classification, but can also be used for regression. SVMs can be used in conjunction with either linear models or kernel-based models (see Section~\ref{kernel_models}). In the linear case, an objective is used which only penalises predictions when they deviate from the training data by more than some threshold~\cite[Equation (12.36)]{hastie2009elements}, namely
\begin{equation}\label{svm}
\bbeta:=\displaystyle\argmin_{\tilde{\bbeta}\in\RR^m}\sum_{i=1}^n V_{\epsilon}(y_i-\bx_i^{\top}\tilde{\bbeta})+\frac{\sigma^2}{2}\|\tilde{\bbeta}\|_2^2,
\end{equation}
where
$$V_{\epsilon}(r):=\left\{\begin{array}{ll}0&|r|<\epsilon\\|r|-\epsilon&\textrm{otherwise.}\end{array}\right.$$
The solution to (\ref{svm}) typically has the distinctive property that it is a linear combination of a subset $S\subset\{1,\ldots,n\}$ of the training samples, known as the support vectors~\cite[Chapter 12]{hastie2009elements}, giving
$$\bbeta:=\sum_{i\in S}s_i\,\bx_i,$$
for some $\{s_i\}$. There is no closed-form solution for the $\{s_i\}$, but the solution to the convex problem (\ref{svm}) can be obtained using efficient algorithms.

\subsection{Uncertainty propagation results}\label{linear_results}

Recall that linear regression, penalised linear regression and linear support vector machines all result in a linear model of the form 
\begin{equation}\label{linear_model}
y^{\ast}=\bbeta^{\top}\bx^{\ast}.
\end{equation}
In this section, we present analytical expressions for the mean and variance of this point prediction with respect to input variable uncertainty. The results in this section are standard results in multivariate statistics and we claim no novelty but reproduce them here for completeness. 

Suppose then that the vector of input variables for a given test point $\bX^{\ast}$ has known mean $\bmu$ and known covariance matrix $\bGam$. We note that $\bmu$ and $\bGam$ can be computed given the probability density function (pdf) $f(\bx)$ of $\bX$ if necessary. We have the following expressions for the mean and variance of the point predictions.

\begin{thm}\label{general_linear}
Let $\bX$ be such that $\EE\,\bX=\bmu$ and $\mathrm{Cov}\,\bX=\bGam$ and suppose a prediction $Y^{\ast}$ is obtained from the linear model (\ref{linear_model}). Then
$$\begin{array}{l}
\EE\,Y^{\ast}=\bbeta^{\top}\bmu;\\
\mathrm{Var}\,Y^{\ast}=\bbeta^{\top}\bGam\bbeta.
\end{array}$$
\end{thm}

Theorem~\ref{general_linear} is a standard result and can be found for example in~\cite[Equation 2-43]{johnson2002applied}.

More can be said in the specific case that $\bX^{\ast}$ follows a multivariate Gaussian distribution. In this case, the prediction is also a (univariate) Gaussian random variable with mean and variance corresponding to those given in Theorem~\ref{general_linear}.

\begin{thm}\label{gaussian_linear}
Let $\bX^{\ast}\sim\mathrm{N}(\bmu,\bGam)$ and suppose a prediction $Y^{\ast}$ is obtained from the linear model (\ref{linear_model}). Then
$$Y^{\ast}\sim\mathrm{N}(\bbeta^{\top}\bmu,\bbeta^{\top}\bGam\bbeta).$$
\end{thm}

Theorem~\ref{general_linear} is also a standard result and can be found for example in~\cite[Result 4.2]{johnson2002applied}.

We remark that, in the Gaussian case of Theorem~\ref{gaussian_linear}, we have complete knowledge of the pdf of $Y^{\ast}$, which allows credible intervals to be analytically computed. This situation is, however, the exception, and for all other combinations of ML model and input data distribution considered in this paper, results are limited to the moments of $Y^{\ast}$ and the precise distribution of $Y^{\ast}$ is not obtained.

\section{Kernel-based models}\label{kernel_models}

In this section, we characterise the output uncertainty of point predictions from four different kernel-based ML models: kernel ridge regression, Gaussian Processes (GPs), kernel support vector machines (SVMs) and relevance vector machines (RVMs). These characterisations take the form of precise analytical expressions for the mean and variance of the output for certain choices of kernel and certain types of input data distribution $\mD$. 

Two of these methods, GPs and RVMs, are based upon Bayesian inference, and thus return analytical expressions for the posterior distribution of the output variable given the training data. This posterior distribution captures model uncertainty, and its corresponding variance is often used to characterise this uncertainty. If the model is evaluated on a new test point, however, this variance does not capture uncertainty in the input data. 

GPs and RVMs can also be used to make a point prediction, by taking the \emph{maximum a posteriori} (MAP) estimate based on the posterior distribution. In keeping with the framework of uncertainty propagation through fixed ML models, this paper considers the point predictions made by GPs and RVMs and how input uncertainties propagate through these point predictions.

In its most common implementation, kernel ridge regression is viewed as a point predictor. We note, however, that there does exist a Bayesian interpretation of kernel ridge regression which turns out to be equivalent to a GP~\cite[Section 2.1]{williams2006gaussian}.

In Section~\ref{kernel_framework}, we describe the general framework of kernel-based models. In Section~\ref{kernel_background}, we then briefly review the four specific methods before presenting our uncertainty propagation results in Section~\ref{kernel_results}.

\subsection{The general framework}\label{kernel_framework}

Linear models, as considered in Section~\ref{linear_models}, often impose a restrictive assumption, and so an appealing extension is to transform the input data (nonlinearly) into some different feature space. Write $\phi:\RR^m\rightarrow\RR^p$ for this transformation, and write $\bz_i\in\RR^p$ for the transformed input data for observation $i$ ($i=1,\ldots,n$), so that $\bz_i=\phi(\bx_i)$. We write $\bD_{\bx}$ for the $n\times p$ \emph{transformed data matrix} whose rows are the input data observations in feature space. The dimension $p$ of the transformed data need not be equal to the dimension $m$ of the original data, and indeed may actually be infinite.

A \emph{kernel function} $k:\RR^n\times\RR^n\rightarrow\RR$ is a function which computes inner products in the transformed space, that is $k(\bx,\tilde{\bx}):=\phi(\bx)^{\top}\phi(\tilde{\bx})=\bz^{\top}\tilde{\bz}$, without explicitly evaluating the feature vectors $\phi(\bx)$ and $\phi(\tilde{\bx})$ in the transformed space. The ability to perform such inner products without explicitly transforming the data often leads to huge improvements in computational efficiency and scaleability, and is often referred to as the \emph{kernel trick}~\cite{williams2006gaussian}.

We also define some quantities which capture evaluations of the kernel function upon the training set and a new test point. We define a \emph{kernel matrix} $\bK\in\RR^{n\times n}$ whose entries are given by $K_{ij}=k(\bx_i,\bx_j)$, $i,j=1,\ldots,n$. Furthermore, if $\phi(\bx^{\ast})$ is the transformation of a new test point $\bx^{\ast}$ into feature space, we define a vector $\bk^{\ast}\in\RR^n$ whose entries are $k^{\ast}_i:=k(\bx^{\ast},\bx_i)$, $i=1,\ldots,n$.

\subsection{Background on four kernel-based methods}\label{kernel_background}

\subsubsection{Kernel ridge regression}\label{kernel_ridge}

Ridge regression was reviewed in Section~\ref{linear_review} in the context of linear models. Kernel ridge regression refers to performing linear ridge regression in the transformed feature space~\cite{welling2013kernel} as described in Section~\ref{kernel_framework}, which means learning a model of the form 
\begin{equation}\label{kernel_model}
y=\bbeta^{\top}[\phi(\bx)]
\end{equation}
by solving
\begin{equation}\label{kernel_ridge_problem}
\bbeta:=\displaystyle\argmin_{\tilde{\bbeta}\in\RR^m}\|\by-\bD_{\bx}\tilde{\bbeta}\|_2^2+\sigma^2\|\tilde{\bbeta}\|_2^2.
\end{equation}

Analogously to the linear case described in Section~\ref{linear_review}, it is straightforward to show that (\ref{kernel_ridge_problem}) has the closed-form solution~\cite[Equation (5)]{welling2013kernel}
\begin{equation}\label{kernel_solution}
\bbeta=\bD^{\top}_{\bx}(\bD_{\bx}\bD_{\bx}^{\top}+\sigma^2 \tensor{I})^{-1}\by.
\end{equation}

Combining (\ref{kernel_model}) and (\ref{kernel_solution}), it follows that the prediction of the kernel ridge regression model on a new test input $\bx^{\ast}$ is given by
\begin{equation}\label{kernel_prediction}
y^{\ast}=\left[\phi(\bx^{\ast})\right]^{\top}\bD_{\bx}^{\top}(\bD_{\bx}\bD_{\bx}^{\top}+\sigma^2 \tensor{I})^{-1}\by.
\end{equation}

The kernel trick, as described in Section~\ref{kernel_framework}, can then be employed to give an alternative expression for the prediction which does not require explicit transformation of the data. Substituting the definitions of $\bK$ and $\bk^{\ast}$ into (\ref{kernel_prediction}), we obtain
\begin{equation}\label{kernel_regression_point}
y^{\ast}=\left(\bk^{\ast}\right)^{\top}(\bK+\sigma^2 \tensor{I})^{-1}\by.
\end{equation}

\subsubsection{Gaussian Processes}\label{GP}

A Gaussian Process (GP) is a collection of functions with the property that any finite collection of them follows a multivariate Gaussian distribution. Bayesian inference can be performed over a GP by specifying a prior on the collection of predictions~\cite[Section 2.2]{williams2006gaussian}. One standard way in which this is done is to specify a prior of the form $y=f(\bx)+\epsilon$, where $f$ is a GP and where $\epsilon$ is i.i.d. additive Gaussian noise, such that
\begin{equation}\label{GP_prior}\begin{bmatrix}\by\\y^{\ast}\end{bmatrix}\sim \mathrm{N}\left(\mathbf{0},\begin{bmatrix}\bK+\sigma^2\tensor{I}&(\bk^{\ast})^{\top}\\\bk^{\ast}&k(\bx^{\ast},\bx^{\ast})\end{bmatrix}\right).\end{equation}
In (\ref{GP_prior}), $\bK$ and $\bk^{\ast}$ are as defined in Section~\ref{kernel_framework}, and $\sigma^2$ is the variance of i.i.d. Gaussian noise in the prior. Recall that $\by$ denotes the training data and $y^{\ast}$ denotes a new prediction, and that both $\bK$ and $\bk^{\ast}$ are functions of the input variables.

Using a standard result for conditional distributions of multivariate Gaussians, we immediately obtain
\begin{equation}\label{GP_posterior}y^{\ast}|\by\sim\mathrm{N}\left[(\bk^{\ast})^{\top}(\bK+\sigma^2\tensor{I})^{-1}\by,k(\bx^{\ast},\bx^{\ast})-(\bk^{\ast})^{\top}(\bK+\sigma^2\tensor{I})^{-1}\bk^{\ast}\right]
\end{equation}
for the posterior distribution of $Y^{\ast}$ given the training data. A point prediction for such a GP is given by the mean of this distribution, that is
\begin{equation}\label{GP_point}
y^{\ast}=(\bk^{\ast})^{\top}(\bK+\sigma^2\tensor{I})^{-1}\by.
\end{equation}
We note that, for the same choice of $\sigma^2$, the point predictions (\ref{kernel_regression_point}) and (\ref{GP_point}) are identical.

\subsubsection{Kernel support vector machines}

Just as in the case of ridge regression, an SVM can also be solved in a transformed feature space by replacing the data matrix $\bC_{\bx}$ with the transformed data matrix $\bD_{\bx}$ in (\ref{svm}). We omit details, but it turns out that the objective function of the Lagrangian dual to (\ref{svm}) in the transformed space can be expressed in terms of inner products in transformed space, so that the kernel trick can again be employed~\cite[Section 12.3.7]{hastie2009elements}. This time the prediction takes the form 
$$y^{\ast}:=\sum_{i\in S}s_i\,k(\bx^{\ast},\bx_i),$$
for some $\{s_i\}$, where $S$ is a subset of $\{1,\ldots,n\}$ as before. Using the notation introduced in Section~\ref{kernel_ridge}, we can write this point prediction in the form 
\begin{equation}\label{SVM_point}
y^{\ast}=\bal^{\top}\bk^{\ast},\end{equation}
for some $\bal\in\RR^n$.

\subsubsection{Relevance vector machines}\label{RVM}

A relevance vector machine (RVM) model~\cite{tipping2001sparse} takes the same functional form as a support vector machine, that is
$$y=\sum_{i=1}^n w_i^{\top}\phi_i(\bx)+\epsilon,$$
where the $\{\phi_i(\cdot)\}$ are arbitrary basis functions and where $\epsilon$ is assumed to be i.i.d. Gaussian noise with variance $\sigma^2$. We write $\bphi(\bx)$ for the vector of basis functions evaluated at $\bx$. In keeping with the rest of the paper, we assume that all data has been centred and so we consider a simplification of the model proposed in~\cite{tipping2001sparse} in which the basis function which is everywhere equal to $1$ is omitted.

Where the RVM approach differs from that of SVM is that a prior is placed over the weights and Bayesian inference is performed. Writing $\bw=[w_1\;\cdots\;w_n]^{\top}$ for the vector of weights, a standard approach is to assume a prior of $\bw\sim\mathrm{N}(\mathbf{0},\bA^{-1})$, where $\bA$ is typically a diagonal matrix. Given a new input data point $\bx^{\ast}$, we write $\bphi^{\ast}:=\bphi(\bx^{\ast})$. An application of Bayes' rule leads to a posterior distribution of
$$\bw\sim\mathrm{N}\left[\frac{1}{\sigma^2}\bSigma\bPhi^{\top}\by,(\bphi^{\ast})^{\top}\bSigma^{-1}\bphi^{\ast}\right],$$
where $\bPhi$ has entries $\Phi_{pq}=\phi_q(\bx_p)$ and where
$$\bSigma:=\left(\frac{1}{\sigma^2}\bPhi^{\top}\bPhi+\bA\right)^{-1}.
$$
A point prediction for $y^{\ast}$ is then given by
$$y^{\ast}=\frac{1}{\sigma^2}(\bphi^{\ast})^{\top}\bSigma\bPhi^{\top}\by.$$
If the basis functions are chosen so that $\bphi_i(\bx)=k(\bx_i,\bx)$, the kernel trick can again be employed, and it can be shown~\cite{quinonero2003prediction} that the point prediction can also be written in the form
\begin{equation}\label{RVM_point}
y^{\ast}=\bal^{\top}\bk^{\ast},
\end{equation}
for some $\bal\in\RR^n$.

\subsection{Uncertainty propagation results}\label{kernel_results}

Reviewing the point prediction expressions in Section~\ref{kernel_background}, namely (\ref{kernel_regression_point}), (\ref{GP_point}), (\ref{SVM_point}) and (\ref{RVM_point}), we see that they all take the form
\begin{equation}\label{kernel_point}
y^{\ast}=\bal^{\top}\bk^{\ast},
\end{equation}
for some $\bal\in\RR^n$. Or to put it another way, each of the point predictions is linear in the transformed kernel space. We also see 
that the point predictions for two of the methods, kernel ridge regression and GPs, are identical when the same choices of kernel and regularisation/noise parameter are made. 

In this section, we obtain analytical expressions for the mean and variance of point predictions of the general form (\ref{kernel_point}) with respect to input variable uncertainty. These general results can then be used to propagate uncertainties analytically through any of the four models by making the appropriate substitution for $\bal$. Our results and method of proof are inspired by those presented in~\cite{quinonero2003prediction}. Proofs of all  novel  results presented in this section are given in Appendix~\ref{kernel_proofs}.

Suppose then that the vector of input variables for a given test point $\bX^{\ast}$ has pdf $f(\bx)$. We introduce some new notation. Write $\bl\in\RR^n$ for the mean of $\bk^{\ast}$, that is a vector whose $i^\textrm{th}$ component is given by
\begin{equation}\label{l_def}
l_i:=\int k(\bx,\bx_i)f(\bx)\textrm{d}\bx,
\end{equation}
and write $\bL\in\RR^{n\times n}$ for the matrix whose $(i,j)^\textrm{th}$ component is given by
\begin{equation}\label{L_def}
L_{ij}:=\int k(\bx,\bx_i)k(\bx,\bx_j)f(\bx)\textrm{d}\bx.
\end{equation}

We have the following general expressions for the mean and variance of the point predictions.

\begin{thm}\label{general_kernel}
$$\begin{array}{l}
\EE\,Y^{\ast}=\bal^{\top}\bl;\\
\mathrm{Var}\,Y^{\ast}=\bal^{\top}\bL\bal-(\bal^{\top}\bl)^2.
\end{array}$$
\end{thm}

Theorem~\ref{general_kernel} is a simplification of results derived in~\cite[Section 4.2.2]{quinonero2003prediction}  for GPs and RVMs , in which we ignore model uncertainty contributions.  It also extends the results in~\cite[Section 4.2.2]{quinonero2003prediction} to any kernel-based model as defined in (\ref{kernel_point}). 

We next present the specific form that Theorem~\ref{general_kernel} takes in the case of the popular radial basis function (RBF) kernel and some common input data distributions. We first obtain a result for the case where the input data distribution is a multivariate Gaussian distribution. Then we also obtain results for the case where the input variables are independent uniform or symmetric triangular distributions.

\subsubsection{Specific results for RBF kernels}

We define the RBF kernel in the conventional way, such that
\begin{equation}\label{RBF}
k(\bx,\tilde{\bx}):=\mathrm{exp}\left[-\frac{1}{2}(\bx-\tilde{\bx})^{\top}\bLam^{-1}(\bx-\tilde{\bx})\right],
\end{equation}
where $\bLam=\mathrm{diag}(\lambda_1^2,\ldots,\lambda_m^2)$, allowing for different length scales in different input variables.

First suppose that $\bX^{\ast}$ follows a multivariate Gaussian distribution, $\bX^{\ast}\sim\mathrm{N}(\bmu,\bGam)$. Defining $\bm_{ij}:=\frac{1}{2}(\bx_i+\bx_j)$ to be the midpoint between two training input vectors $\bx_i$ and $\bx_j$, the following result was derived in~\cite{candela2003propagation,quinonero2003prediction}

\begin{thm}[{\cite[Equations (33) and (37)]{quinonero2003prediction}}]\label{kernel_mvg}
Let $\bX^{\ast}\sim\mathrm{N}(\bmu,\bGam)$ and suppose a point prediction is obtained using the RBF kernel given in (\ref{RBF}). Then Theorem~\ref{general_kernel} holds with
\begin{equation}\label{mvg_l}
l_i:=|\bLam^{-1}\bGam+\tensor{I}|^{-\frac{1}{2}}\exp\left[-\frac{1}{2}(\bmu-\bx_i)^{\top}(\bLam+\bGam)^{-1}(\bmu-\bx_i)\right];
\end{equation}
\begin{equation}\label{mvg_L}
L_{ij}:=|2\bLam^{-1}\bGam+\tensor{I}|^{-\frac{1}{2}}\exp\left(-\frac{1}{2}\left[(\bmu-\bm_{ij})^{\top}\left(\frac{\bLam}{2}+\bGam\right)^{-1}(\bmu-\bm_{ij})+(\bx_i-\bx_j)^{\top}\left(2\bLam\right)^{-1}(\bx_i-\bx_j)\right]\right).
\end{equation}
\end{thm}

It can be shown that, in the limit as $\bGam\rightarrow\mathbf{0}$, we have $l_i\rightarrow k(\bmu,\bx_i)$ and 
$L_{ij}\rightarrow k(\bmu,\bx_i)k(\bmu,\bx_j)$, so that $\bL\rightarrow \bl\,\bl^{\top}$~\cite{quinonero2003prediction}. Substituting into Theorem~\ref{general_kernel} we see that $\EE\,Y^{\ast}$ tends to the point prediction at $\bX^{\ast}=\bmu$, with zero variance, as expected. 

We can also obtain results in the case where the components of $\bX^{\ast}$ are independent. This turns out to be possible due to the separable nature of the RBF kernel, which can be written as
\begin{equation}\label{separable_RBF}
k(\bx,\tilde{\bx}):=\prod_{p=1}^m\exp\left[-\frac{(x^p-\tilde{x}^p)^2}{2\lambda_p^2}\right],
\end{equation}
where here $x^p$ denotes the $p^{\mathrm{th}}$ component of $\bx$.

Now if the components of $\bX^{\ast}$ are independent, the pdf of $\bX^{\ast}$, $f(\bx)$, is also separable, and can be written
\begin{equation}\label{separable_pdf}
f(\bx)=\prod_{p=1}^m f^p(x^p).
\end{equation}
Denoting by $x_i^p$ the $p^{\mathrm{th}}$ component of the training input vector, the following characterisation of $\bl$ and $\bL$ was given in~\cite{wan2014analytical}.
\begin{thm}[{\cite[Equations (20) and (21)]{wan2014analytical}}]\label{kernel_ind}
Let $\bX^{\ast}$ have independent components with pdf as in (\ref{separable_pdf}) and suppose a point prediction is obtained using the RBF kernel given in (\ref{RBF}). Then Theorem~\ref{general_kernel} holds with
$$l_i:=\prod_{p=1}^m l_i^p\;\;\;\;\textrm{and}\;\;\;\;L_{ij}:=\prod_{p=1}^m L_{ij}^p,\;\;\;\;\textrm{where}$$
\begin{equation}\label{separable_l_def}
l_i^p:=\int \exp\left[-\frac{(x^p-x_i^p)^2}{2\lambda_p^2}\right]f^p(x^p)\textrm{d}x^p
\end{equation}
and 
\begin{equation}\label{separable_L_def}
L_{ij}:=\int \exp\left[-\frac{(x^p-x_i^p)^2}{2\lambda_p^2}\right]\exp\left[-\frac{(x^p-x_j^p)^2}{2\lambda_p^2}\right]f(x^p)\textrm{d}x^p.
\end{equation}
\end{thm}

 The result in~\cite{wan2014analytical} includes an extra multiplicative factor relating to a GP hyperparameter, and so we include a proof for this result in Appendix~\ref{kernel_proofs} for completeness.

 We next give expressions for the specific form that $l_i^p$ and $L_{ij}^p$ take for two distributions that often arise in uncertainty analysis.  A result was obtained for the uniform distribution in~\cite{wan2014analytical}, and we also obtain a novel result for the symmetric triangular distribution. We denote by $\Phi(x)$ the cumulative distribution function for the standard univariate Gaussian distribution $N(0,1)$, and we write $m_{ij}^p$ for the $p^{\mathrm{th}}$ component of $\bm_{ij}$.

\begin{thm}[{\cite[page 14]{wan2014analytical}}]\label{kernel_uniform}
Let the  $p^{\mathrm{th}}$ component of $\bX^{\ast}$ be uniformly distributed on the interval $[a,b]$. Then
\begin{equation}\label{l_uniform}l_i^p:=\frac{\lambda_p\sqrt{2\pi}}{b-a}\left[\Phi\left(\frac{b-x_i^p}{\lambda_p}\right)-\Phi\left(\frac{a-x_i^p}{\lambda_p}\right)\right]
\end{equation}
and
\begin{equation}\label{L_uniform}L_{ij}^p:=\frac{\lambda_p\sqrt{\pi}}{b-a}\exp\left[-\frac{(x_i^p-x_j^p)^2}{4\lambda_p^2}\right]\left[\Phi\left(\frac{b-m_{ij}^p}{(\lambda_p/\sqrt{2})}\right)-\Phi\left(\frac{a-m_{ij}^p}{(\lambda_p/\sqrt{2})}\right)\right].\end{equation}
\end{thm}

 The result in~\cite{wan2014analytical} includes an extra multiplicative factor relating to a GP hyperparameter, and so we also include a proof for this result in Appendix~\ref{kernel_proofs} for completeness.

\begin{thm}\label{kernel_tri}
Let the  $p^{\mathrm{th}}$ component of $\bX^{\ast}$ follow a symmetric triangular distribution on the interval $[a,b]$. Then
\begin{equation}\label{l_triangular}
l_i^p:=\begin{array}{l}\frac{4\lambda_p^2}{(b-a)^2}\left\{\exp\left[-\frac{(a-x_i^p)^2}{2\lambda_p^2}\right]+\exp\left[-\frac{(b-x_i^p)^2}{2\lambda_p^2}\right]-2\exp\left[-\frac{(\frac{a+b}{2}-x_i^p)^2}{2\lambda_p^2}\right]\right.\\
+\left.\frac{\sqrt{2\pi}}{\lambda_p}\left[(a-x_i^p)\Phi\left(\frac{a-x_i^p}{\lambda_p}\right)+(b-x_i^p)\Phi\left(\frac{b-x_i^p}{\lambda_p}\right)-(a+b-2x_i^p)\Phi\left(\frac{\frac{a+b}{2}-x_i^p}{\lambda_p}\right)\right]\right\}
\end{array}
\end{equation}
and
\begin{equation}\label{L_triangular}
L_{ij}^p:=\begin{array}{l}\frac{2\lambda_p^2}{(b-a)^2}\exp\left[-\frac{(x_i^p-x_j^p)^2}{4\lambda_p^2}\right]\left\{\exp\left[-\frac{(a-m_{ij}^p)^2}{\lambda_p^2}\right]+\exp\left[-\frac{(b-m_{ij}^p)^2}{\lambda_p^2}\right]-2\exp\left[-\frac{(\frac{a+b}{2}-m_{ij}^p)^2}{\lambda_p^2}\right]\right.\\
+\left.\frac{2\sqrt{\pi}}{\lambda_p}\left[(a-m_{ij}^p)\Phi\left(\frac{a-m_{ij}^p}{(\lambda_p/\sqrt{2})}\right)+(b-m_{ij}^p)\Phi\left(\frac{b-m_{ij}^p}{(\lambda_p/\sqrt{2})}\right)-(a+b-2m_{ij}^p)\Phi\left(\frac{\frac{a+b}{2}-m_{ij}^p}{(\lambda_p/\sqrt{2})}\right)\right]\right\}.\end{array}
\end{equation}
\end{thm}

We have derived results for independent input variables in the cases where each input feature follows a uniform or symmetric triangular distribution. We note that it would be possible to derive similar results for any other univariate distribution $f(x^p)$ for which the integrals (\ref{separable_l_def}) and (\ref{separable_L_def}) are tractable.

\section{Numerical experiments}\label{experiments}

\subsection{State-of-health estimation for lithium-ion cells}\label{batteries}

We illustrate in this paper how our method can be used in the context of a specific metrology case study: state-of-health estimation for lithium-ion batteries. In this section we give background on this case study, summarising the description originally presented in~\cite{chan2022comparison}. 

Demand for lithium-ion cells is growing rapidly, in large part due to the expanding electric vehicle market~\cite{kotak2021end}. However lithium-ion cells age due to repeated charge/discharge cycles and during storage. Once capacity is reduced to below operational limits, typically 70–80 \% of the original capacity, a lithium-ion cell needs to be replaced. Aged lithium-ion cells can still be useful for lower power- or energy-density applications, such as domestic energy storage. Such cells are called second-use cells or second-life cells~\cite{shahjalal2022review}. By repurposing aged cells, the demand for raw materials to manufacture new cells can be reduced and the cost of second-use applications can be decreased. 

For the economically viable application of second-use cells, an accurate and cost-effective method for estimating state-of-health (SOH) is essential. SOH refers to the actual charge capacity that can be stored in a cell, as a proportion of the nominal capacity value. 

The most common method to determine SOH is by integrating the transient current during one complete charge/discharge cycle at a nominal operating
current. However this method is time-consuming, which motivates the use of ML  approaches in which an empirical model that maps a defined set of ageing parameters to SOH is learned from data. This case study concerns the use of Electrochemical Impedance Spectroscopy (EIS) measurements combined with equivalent circuit models as a data source for SOH estimation. In EIS, the impedance of a chemical system is measured at multiple angular frequencies, so that an impedance spectrum is obtained.

It is also important to understand the reliability of any SOH estimation method, which includes obtaining a quantitative assessment of the uncertainty of the estimates. Such uncertainty quantification can be used, for example, to ensure reliable decisions are taken in conformity assessment. Uncertainty arises from two sources. Firstly, there is uncertainty concerning the data used both for training the model and as inputs for a new test measurement. Data uncertainty is caused by both measurement uncertainty and cell-to-cell variation in the EIS spectrum corresponding to a particular SOH value. Secondly, there exists uncertainty in the choice and fitting of the model describing the relationship between the ageing parameters and the SOH. These two sources of uncertainty correspond to the notions of data and model uncertainty discussed in Section~\ref{sec:intro}.

In Section~\ref{sec:data}, we describe a dataset consisting of SOH and EIS measurements. We outline how these measurements were obtained and briefly describe an equivalent circuit model to preprocess the EIS data. In Section~\ref{sec:EISunc}, we describe how EIS data uncertainties were modelled and propagated through to equivalent circuit coefficient uncertainties in~\cite{chan2022comparison}, which then become input uncertainties for our own method. 

\subsubsection{EIS and SOH data}\label{sec:data}

We summarise the data collection described in~\cite[Section 2]{chan2022comparison} and refer the interested reader there for more comprehensive details. Life-cycle tests (LCTs) were conducted by four different measurement institutions on lithium-ion cells obtained from a commercial supplier. SOH was computed by taking the ratio of the discharge capacity of the aged cell to the initially measured capacity of the fresh cell. Following each capacity measurement, EIS impedance spectra were also measured. A public release of this dataset is available at~\cite{chan2002release}. The dataset was reduced to only include samples for which the SOH is above 75 \%, which resulted in a dataset consisting of 165 samples.

Following inspection of the impedance spectra, all data were fitted to
a so-called 2ZARC equivalent circuit whose impedance for a given angular frequency has an analytical expression in terms of 12 coefficients comprising resistances, inductances, characteristic angular frequencies and phase angles. The 2ZARC equivalent circuit model is described in detail in~\cite[Section 3.2]{chan2022comparison}. See~\cite[Appendix A]{chan2022comparison} for full details of the method for fitting the equivalent circuit model. 

A polynomial regression model of order two (quadratic) for state-of-health was fitted in~\cite{chan2022comparison}, and stepwise linear regression was used to identify which variables to include in the model. Following this approach, we discovered that it is sufficient to use only four variables in the model: two of the 12 coefficients for prediction of state-of-health along with their squared values. More specifically, these two coefficients represent a resistance and a capacitance in the equivalent circuit, and in keeping with~\cite{chan2022comparison} we will refer to them henceforth as \texttt{Rct2} and \texttt{Cdl2}.

\subsubsection{Uncertainty of equivalent circuit coefficients due to temperature fluctuation}\label{sec:EISunc}

Three potential sources of data uncertainty in the raw EIS data were identified in~\cite{chan2022comparison}, namely temperature (inhomogeneities in the temperature distribution in the measurement chamber and in the cell), measurement time (cells never fully equilibrate) and the calibration of the impedance meter. In~\cite{chan2022comparison}, uncertainties due to one of these sources, temperature fluctuations, were propagated through the equivalent circuit model using Monte Carlo sampling as follows.

Writing $Z$ for impedance, $f$ for angular frequency and $T$ for temperature, the sensitivity coefficient $\partial Z(T,f)/\partial T$ for impedance with respect to temperature fluctuation was assessed by conducting EIS
measurements at three different controlled temperatures. The temperature sensitivity coefficient was modelled as a function
of frequency by fitting a linear regression to the impedance data $Z(T,f)$ across the three temperature values, as
$$Z(T,f)\approx Z(T_{\textrm{ref}},f)+\frac{\partial Z(T,f)}{\partial T}(T-T_{\textrm{ref}}),$$
where $T_{\textrm{ref}}$ is a reference temperature.
As temperature sensitivity data was only available at discrete frequency
values, the partial derivative $\partial Z(T,f)/\partial T$ was interpolated to
other angular frequencies where required.

Having determined the sensitivity coefficient, a set of $n$ temperature values $T_i$ was then sampled at random from the defined probability distribution for temperature, which was modelled as a uniform distribution centred on the measurement with a width of $1\;K$. For each original spectrum $Z(T_{\textrm{ref}},f)$, a set of $n$ temperature-distorted spectra was then generated, as
$$Z(T_i,f)=Z(T_{\textrm{ref}},f)+\frac{\partial Z(T_i,f)}{\partial T}(T_i-T_{\textrm{ref}}).$$
Equivalent circuit coefficients were then evaluated for each spectrum $Z(T_i,f)$ using the nonlinear least squares fitting procedure described in Section~\ref{sec:data}. 

We followed the same Monte Carlo sampling approach used in~\cite{chan2022comparison}, using 1000 samples, subsequently evaluating the sample mean and covariance matrix for the equivalent circuit coefficients for each impedance spectrum. 

It should be noted that model uncertainty concerning the equivalent circuit fit is not here taken into account. In addition, only the contribution of temperature fluctuations to the data uncertainty is accounted for in~\cite{chan2022comparison}. However, the work we present here is useful as a demonstration of how known data uncertainties of any kind can be explicitly accounted for in a metrology application, and our method would extend to any other sources of uncertainty that can be characterised. It would be interesting to extend the application of our method to other sources of uncertainty such as measurement time uncertainty and impedance meter calibration, but this would require a new campaign of experiments to be conducted.

\subsubsection{Models}\label{models}

We train two models for predicting state-of-health from the equivalent circuit coefficients described in Section~\ref{sec:data}. These two models are chosen to illustrate each of the two families of models considered in Section~\ref{linear_results} and~\ref{kernel_results} respectively. 

The first model is the same polynomial regression model used in~\cite{chan2022comparison}, which can be viewed as a linear model over four input variables consisting of two equivalent circuit coefficients, \texttt{Rct2} and \texttt{Cdl2}, and their squared values. The model was fitted using the standard ordinary least squares method.

The second model is a GP over the same two equivalent circuit coefficients. Since a GP captures nonlinear effects, there is no benefit in this case in including the squares of the variables. An RBF kernel as defined in (\ref{RBF}) is used.  Recall from Section~\ref{GP} that GPs involve hyperparameters that must be optimised. The hyperparameters in this case are the regularisation parameter $\sigma^2$ and the RBF variance parameter $\lambda^2$. The hyperparameters are optimised by the standard approach of maximising the marginal likelihood.

\subsection{Validation of model accuracy}

The focus of this paper is not on optimising model accuracy, and indeed the results presented in this paper are valid for any model, regardless of the model's accuracy. However, the accuracy of the models used is still valuable context for the later experiments, and so here we briefly present a validation of the two models under consideration.

We use RMSE as a model accuracy metric, and we average over 100 randomly drawn 75:25 train/test splits. Since it is only realistic to expect an ML model to interpolate the training data, any samples outside the range of SOH values in the training set were removed from the test set. The average RMSE for the linear model was 0.074, and the average RMSE for the GP model was 0.055. We note that the GP model gives a more accurate fit than the polynomial regression, and that the RMSE is quite small, especially for the GP model.

Figure~\ref{fit_examples} provides an illustration in the form of a plot of predicted SOH against recorded SOH for one instantiation of the test set for both the linear and the GP model. In this specific case, the RMSE for the linear model was 0.068 and the RMSE for the GP model was 0.046.

\begin{figure}[h!]
     \centering
     \begin{subfigure}[b]{0.45\textwidth}
         \centering
         \includegraphics[width=\textwidth]{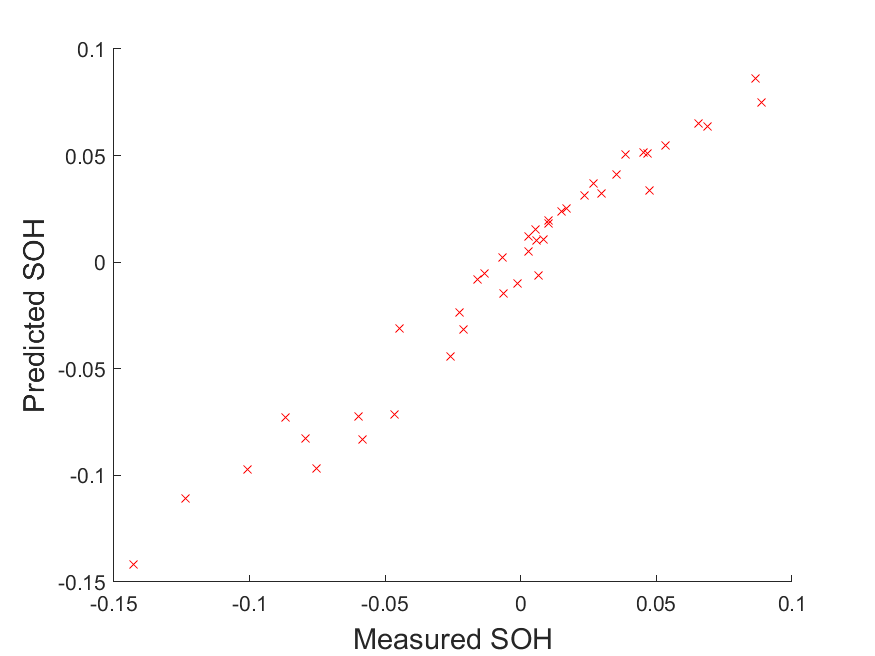}
         \caption{Linear model (RMSE 0.068).}
         \label{linear_sub}
     \end{subfigure}
     \begin{subfigure}[b]{0.45\textwidth}
         \centering
         \includegraphics[width=\textwidth]{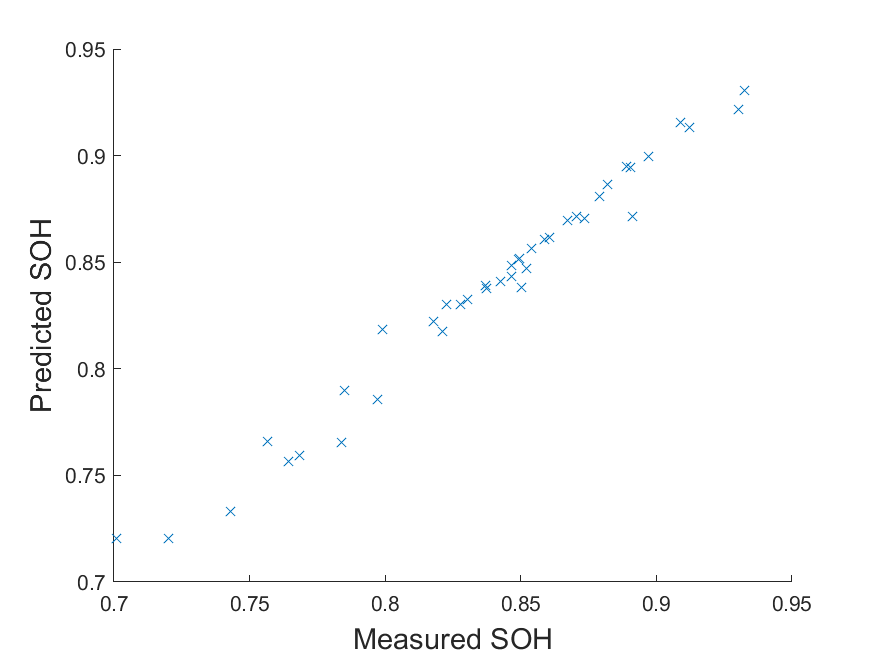}
         \caption{GP model (RMSE 0.046).}
         \label{GP_sub}
     \end{subfigure}
     \caption{Plots of predicted SOH against recorded SOH on one instantiation of the test set.}\label{fit_examples}
\end{figure}

Figure~\ref{surface_plots} shows a visualisation of the linear and GP models of SOH against the two input variables, Rct2 and Cdl2, for the same test set for which results were displayed in Figure~\ref{GP_sub}. The plots are over the range of values of the input variables in the corresponding training set.

\begin{figure}[h!]
     \centering
     \begin{subfigure}[b]{0.45\textwidth}
         \centering
         \includegraphics[width=\textwidth]{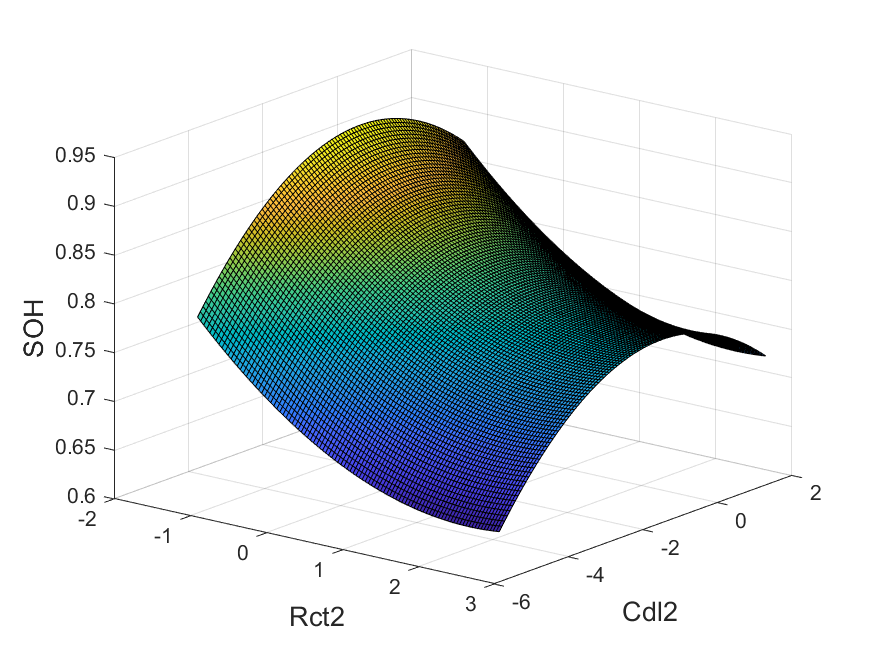}
         \caption{Linear model (RMSE 0.068).}
         \label{linear_surface}
     \end{subfigure}
     \begin{subfigure}[b]{0.45\textwidth}
         \centering
         \includegraphics[width=\textwidth]{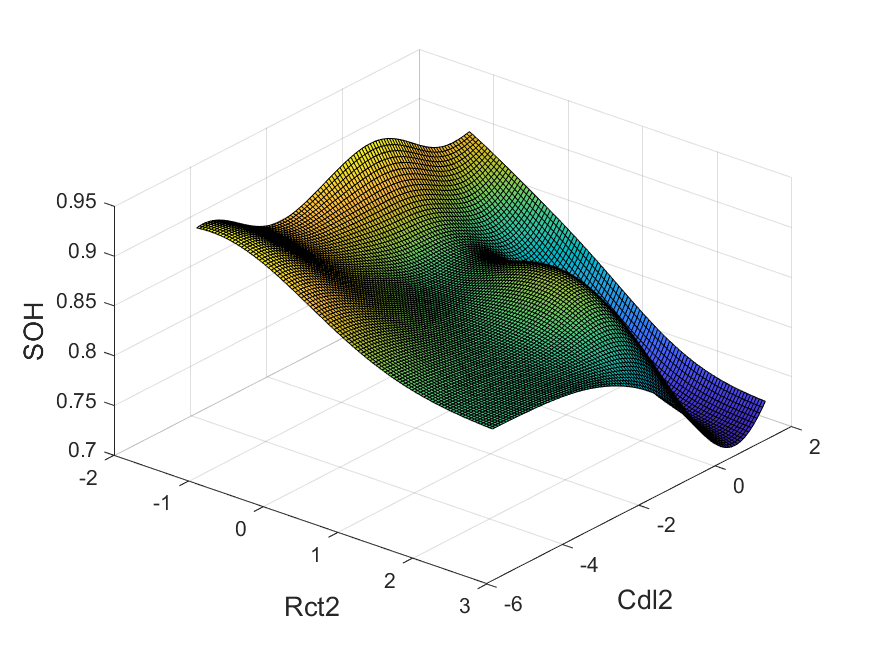}
         \caption{GP model (RMSE 0.046).}
         \label{GP_surface}
     \end{subfigure}
     \caption{Visualisation of linear and GP models for SOH.}\label{surface_plots}
\end{figure}

It is interesting to observe that the more sophisticated GP model appears to be better able to capture some of the nonlinear relationships between input and output variables.

\subsection{Validation of the analytical expressions}\label{accuracy_validation}

In this section, we present evidence that a Monte Carlo sampling approach gives results that are broadly consistent with the evaluation of the expressions given in Sections~\ref{linear_results} and~\ref{kernel_results}. We do this by comparing the returned mean and variance with those returned by a Monte Carlo sampling approach.

For both models described in Section~\ref{models}, a random 75:25 train/test split is used, giving a training set size of 124 and a test set size of 41. The training set input data is centred and normalised, and the same transformation is then applied to the test set input data. Since it is only realistic to expect an ML model to interpolate the training data, one sample outside the range of SOH values in the training set was removed from the test set, leaving 40 samples.

For each sample in the test set, the distribution of the input data (consisting of four variables for the linear regression and two variables for the GP) is modelled as a multivariate Gaussian distribution with mean equal to the recorded value and covariance matrix obtained from the uncertainty propagation technique described in Section~\ref{sec:EISunc}.

For each sample in the test set, the mean and variance of the output of the two ML models is computed both using the analytical expressions from Sections~\ref{linear_results} and~\ref{kernel_results} and using Monte Carlo sampling with number of samples varied over $\{10^3,10^4,10^5,10^6\}$. We measure agreement between the analytical and Monte Carlo computations using RMSE, and we write $\kappa(\EE\,Y^{\ast})$ and $\kappa(\textrm{Var}\,Y^{\ast})$ for the RMSEs for the means and the variances respectively.

Table~\ref{accuracy_table} gives the RMSEs over the 40 test data points for the means and variances. We observe that the RMSE decreases with the number of Monte Carlo trials, and that it is below $10^{-5}$ for all quantities with $10^6$ trials. These results provide strong evidence that the analytical expressions are in agreement with Monte Carlo sampling.

\begin{table}[h!]
    \centering
    \begin{tabular}{|c|cc|cc|}
    \hline
          &\multicolumn{2}{c|}{Linear model} & \multicolumn{2}{c|}{GP}\\
        Number of MC trials &  $\kappa(\EE\,Y^{\ast})\times 10^3$ & $\kappa(\textrm{Var}\,Y^{\ast})\times 10^5$ & $\kappa(\EE\,Y^{\ast})\times 10^3$ & $\kappa(\textrm{Var}\,Y^{\ast})\times 10^6$\\
        \hline
        $10^3$ & 0.1384 & 0.1058 & 0.1024 & 0.6767\\
      $10^4$ & 0.0315 & 0.0294 & 0.0277 & 0.2302\\
      $10^5$ & 0.0131 & 0.0193 & 0.0057 & 0.0305\\
      $10^6$ & 0.0039 & 0.0067 & 0.0025 & 0.0113\\
      \hline
    \end{tabular}
    \caption{RMSE over the test set for varying numbers of Monte Carlo trials for multivariate Gaussian input variables.}
    \label{accuracy_table}
\end{table}

We have provided validation of the analytical results for multivariate Gaussian input distributions (for both linear and kernel-based models). We next provide a similar validation of the results in Theorems~\ref{kernel_uniform} and~\ref{kernel_tri}. To do this, we use the same GP model for state-of-health estimation described earlier in this section and this time model the two input variables as being independently distributed according to either uniform or symmetric triangular distributions which are centred around the best estimate $\bmu$ and with variance equal to the sample variance obtained from the uncertainty propagation technique described in Section~\ref{sec:EISunc}. Given variance $\sigma^2$, the width of the corresponding uniform and symmetric triangular distribution is $\sqrt{12}\sigma$ and $\sqrt{24}\sigma$ respectively~\cite{walck1996hand}.

The same test, comparing the mean and variance obtained from the analytical expressions given in Theorems~\ref{kernel_uniform} and~\ref{kernel_tri} with those obtained from Monte Carlo sampling with varying numbers of samples, was carried out for the  case of independent uniform and symmetric triangular distributions. 

Table~\ref{accuracy_table_ind} gives the RMSEs $\kappa(\EE\,Y^{\ast})$ and $\kappa(\textrm{Var}\,Y^{\ast})$ across all 40 test samples. We observe that the RMSE decreases with the number of Monte Carlo trials, and that it is below $10^{-5}$ for all quantities with $10^6$ trials. These results again provide strong evidence that the analytical expressions are in agreement with Monte Carlo sampling.

\begin{table}[h!]
    \centering
    \begin{tabular}{|c|cc|cc|}
    \hline
          &\multicolumn{2}{c|}{Uniform distribution} & \multicolumn{2}{c|}{Triangular distribution}\\
        Number of MC trials &  $\kappa(\EE\,Y^{\ast})\times 10^4$ & $\kappa(\textrm{Var}\,Y^{\ast})\times 10^6$ & $\kappa(\EE\,Y^{\ast})\times 10^4$ & $\kappa(\textrm{Var}\,Y^{\ast})\times 10^6$\\
        \hline
        $10^3$ & 0.7059 & 0.2164 & 0.5739 & 0.3750\\
      $10^4$ & 0.2752 & 0.0934 & 0.1696 & 0.1265\\
      $10^5$ & 0.0549 & 0.0179 & 0.0571 & 0.0432\\
      $10^6$ & 0.0278 & 0.0063 & 0.0251 & 0.0347\\
      \hline
    \end{tabular}
    \caption{RMSE over the test set for varying numbers of Monte Carlo trials for independent input variables.}
    \label{accuracy_table_ind}
\end{table}

These results also provide evidence that the analytical computations have numerical precision at least to the order of around $10^{-7}$. Analysis of, and methods for further improving, the numerical precision of these computations is left as future work.

\subsection{Computational complexity}\label{complexity}

In this section, we compare the computational efficiency of the analytical and Monte Carlo approaches to uncertainty propagation for both linear and RBF kernel-based models.

\subsubsection{Linear models}\label{linear_complexity}

We begin with the linear case, and start with some intuition. Recalling (\ref{linear_model}), here we consider models of the form $$y^{\ast}=\bbeta^{\top}\bx^{\ast},$$
where $\bbeta\in\RR^m$ and $m$ is the number of input variables.

Suppose a Monte Carlo sampling approach is taken with $T$ trials in the case of a multivariate Gaussian input distribution. Then $mT$ samples from a multivariate Gaussian distribution must be sampled, which involves:
\begin{enumerate}
    \item Drawing $mT$ samples from a standard Gaussian distribution - complexity $\mO(mT)$;
    \item Performing a Cholesky decomposition of the covariance matrix $\bGam$ - complexity $\mO(m^3$);
    \item Multiplying the standard Gaussian samples with the Cholesky factor - complexity $\mO(m^2 T)$;
    \item Performing a scalar product of each of the $T$ length-$m$ input variable samples with the weight vector $\bbeta$ - complexity $\mO(mT)$.
\end{enumerate}

For the analytical approach, recalling Theorem~\ref{general_linear}, a scalar product of length-$m$ vectors and a quadratic form involving an $m\times m$ matrix must be computed. Clearly the computation of the quadratic form has the dominant complexity, which is $\mO(m^2)$.

The above analysis leads us to expect that the Monte Carlo approach scales worse with $m$ than the analytical approach.

The analysis also points to complexity being essentially a function of two variables, $m$ and $T$. In order to empirically confirm the dependence of computational time upon these two variables, we design an experiment involving a synthetic linear model with a varying number of input variables. The model simply computes the mean value of $m$ input variables, that is
$$y^{\ast}=\frac{1}{m}\sum_{l=1}^m x^{\ast}_l,$$
where each input variable takes some value on $[0,1]$. For a given choice of $m$, we consider a multivariate Gaussian input distribution $N(\bmu,\bGam)$, where
$$\bmu:=\frac{1}{2}\mathbf{1}_m;\;\;\;\;\Gamma_{ij}:=\frac{1}{10}\cdot 2^{|i-j|},$$
where here $\mathbf{1}_m$ denotes a vector of $m$ ones. Here the covariance matrix is Toeplitz with covariances decreasing further from the diagonal. We note, however, that the exact form of the input uncertainty is not important here, since the computational complexity depends only on the dimensions of the matrices and vectors involved.

The analytical approach was compared on this test problem with Monte Carlo sampling with numbers of samples varied over $\{10^2,10^3,10^4\}$. The number of input variables was varied over $\{10^2,\lceil 10^{2.5}\rceil,10^3,\lceil 10^{3.5}\rceil,10^4\}$. The results are displayed in Figure~\ref{cube_timing}.

We observe that, as expected, the analytical approach is faster even than the Monte Carlo approach with just $10^2$ samples. Estimating the gradient between $m=10^3$ and $m=10^4$, we observe the expected $\mO(m^2)$ complexity for the analytical method. Perhaps surprisingly, we observe complexity scaling roughly as $m^2$ and sublinearly with $T$ for the Monte Carlo approach. A profiling investigation revealed the dominant operation in this regime to be the matrix-matrix product of the standards Gaussian samples with the Cholesky factor, which from above one would expect to have a naive complexity scaling of $\mO(m^2 T)$. We believe that the improved scaling with respect to $T$ is due to the ability of Matlab to perform vectorised computations sublinearly.

\begin{figure}[h!]
\centering
\includegraphics[width=0.8\textwidth]{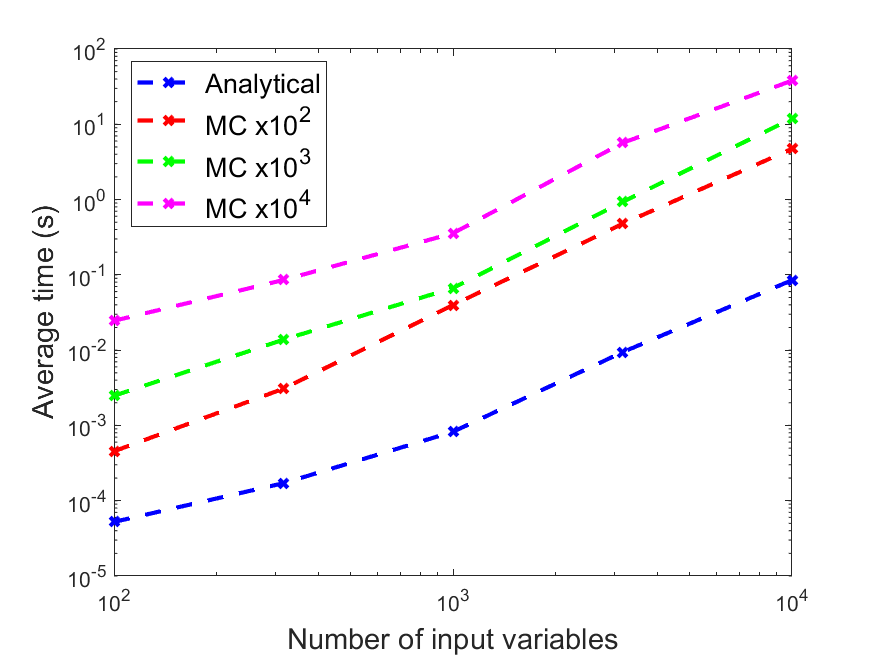}
\caption{Dependence of running time of the analytical and Monte Carlo sampling approaches upon number of input variables and number of Monte Carlo samples for a linear model.}\label{cube_timing}
\end{figure}

\subsubsection{Kernel-based models}

We next turn to investigating the complexity of both analytical and Monte Carlo approaches for kernel-based models.

We again start with some intuition. Suppose first that a Monte Carlo sampling approach is taken with $T$ trials in the case of a multivariate Gaussian input distribution. As well as steps (i) - (iii) described in Section~\ref{linear_complexity}, the input samples must be propagated through the kernel-based point prediction expression (\ref{kernel_point}). This involves $nT$ kernel evaluations, where $n$ is the number of training samples. In the case of the radial basis function, recalling (\ref{RBF}), these kernel evaluations are squared exponentials of scalar products of length-$m$ vectors.

Provided that the number of training samples $n$ is significantly larger than the number of input variables $m$, which would typically be the case where ML is used in metrology, the dominant computational load is likely to come from the squared exponentials, giving a computational complexity of $\mO(nT)$.

For the analytical approach, some $\mO(m^3)$ operations such as matrix inversions and determinant evaluations are required, but under the assumption $m\ll n$ the complexity is again likely to be determined by the number of squared exponential computations. $n$ such computations are required to evaluate $\bl$ using (\ref{l_def}) and $n^2$ such computations are required to evaluate $\bL$ using (\ref{L_def}), giving a computational complexity of $\mO(n^2)$.

The above analysis this time points to complexity being essentially a function of two variables, $n$ and $T$. Furthermore, the analysis would lead us to expect that,  when the number of Monte Carlo trials is increased beyond roughly the number of training samples, the Monte Carlo approach becomes more time consuming. In order to empirically confirm this intuition, we design an experiment involving a GP model over a single input variable with a varying number of training samples.

We use a test problem originating from~\cite{bachstein2019uncertainty} and used subsequently in~\cite{thompson2021uncertainty}, in which synthetic data is generated by the function
$$f(x)=2\cdot\left[\frac{x}{10}+\sin\left(\frac{4x}{10}\right)+\sin\left(\frac{13x}{10}\right)\right],$$
restricted to the interval $[-5,5]$. We consider a homoscedastic noise model in which we add independent Gaussian noise, such that our observational data is given by
$$y_i=f(x_i)+\epsilon_i,\;\;\;\;\epsilon_i\sim\mathcal{N}(0,\sigma^2)$$
for some noise variance $\sigma^2$. 

Training sets of varying sizes are sampled uniformly at random from the interval $[-5,5]$ with $\sigma=1$. A GP with an RBF kernel is then trained, using maximum likelihood to optimise the hyperparameters as described in Section~\ref{models}.

The model for the extreme cases of $n=10^2$ and $n=10^4$ is shown in Figure~\ref{bachstein}(a), with the underlying noiseless model superimposed.

The analytical approach was compared on this test problem with Monte Carlo sampling with numbers of Monte Carlo samples varied over $\{10^2,10^3,10^4\}$. The number of training samples was varied over $\{10^2,\lceil 10^{2.5}\rceil,10^3,\lceil 10^{3.5}\rceil,10^4\}$. The results are displayed in Figure~\ref{bachstein}(b).

\begin{figure}[h!]
     \centering
     \begin{subfigure}[b]{0.45\textwidth}
         \centering
         \includegraphics[width=\textwidth]{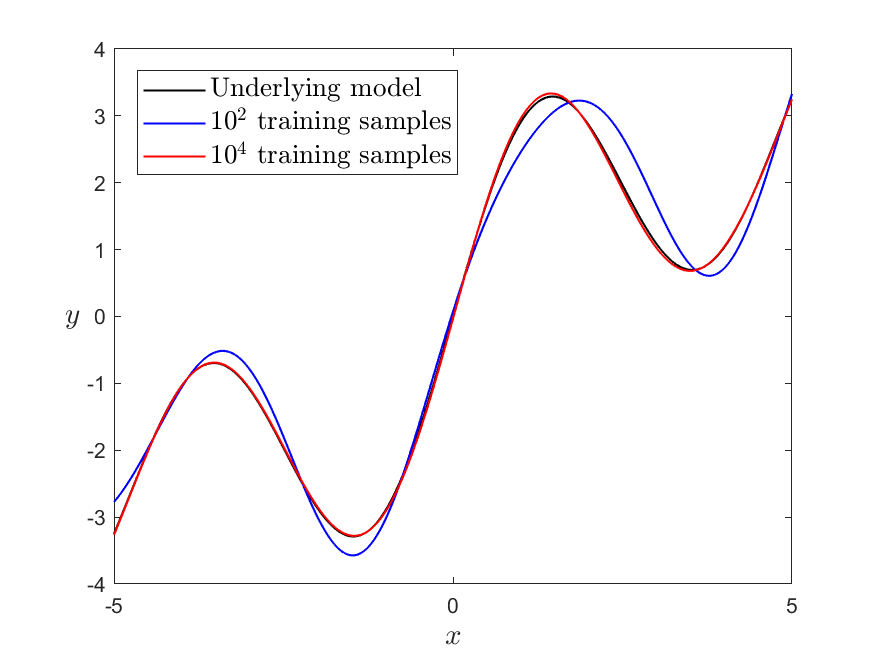}
         \caption{}
     \end{subfigure}
     \begin{subfigure}[b]{0.45\textwidth}
         \centering
         \includegraphics[width=\textwidth]{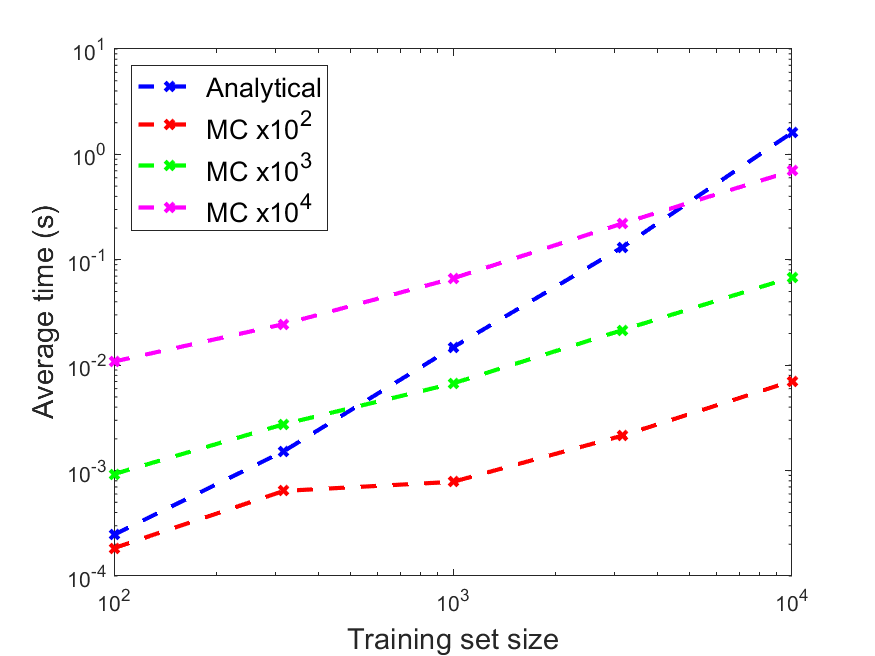}
         \caption{}
     \end{subfigure}
     \caption{(a) GP models for $10^2$ and $10^4$ training samples; (b) Dependence of running time of the analytical and Monte Carlo sampling approaches upon number of training samples and number of Monte Carlo samples for a GP model.}
     \label{bachstein}
\end{figure}

We observe that the numerical results show a striking match to our intuition. Examining the gradients of the log-log plots, we observe that the running time is roughly proportional to $n$ for the Monte Carlo approach and roughly proportional to $n^2$ for the analytical approach. This means that, for a fixed training set size, there is a threshold for the number of Monte Carlo samples beyond which the Monte Carlo method becomes more computationally expensive.

In practice, the number of Monte Carlo samples must be chosen carefully to ensure that the method returns estimates which are suitably accurate. Another way to explain the complexity tradeoff here is that, for a given model, the Monte Carlo method is able to obtain an estimate of the mean and variance of the model to within some accuracy threshold with the same computational expense, but it becomes more computationally expensive if it is desired to improve the accuracy beyond this threshold. 

Determining such thresholds is a nontrivial task, and one of the disadvantages of Monte Carlo approaches is that the accuracy cannot be known \textit{a priori}. Supplement 1 to the GUM~\cite{bipm2008supplement} prescribes an adaptive Monte Carlo method for choosing the number of samples in a way which controls the accuracy. More sophisticated approaches based upon Stein's method have also been proposed~\cite{wubbeler2010two}. Despite these advances, there remains the drawback for any Monte Carlo approach that estimates are subject to non-negligible sampling approximation error.

\section{Conclusions}\label{discussion}

ML models are now widely used in metrology applications, and ensuring the quality and traceability of ML models through uncertainty quantification is an important current challenge for the metrology community. We have chosen in this paper to focus on the challenge of propagation of uncertainty through fixed/trained ML regression models since it helps in isolating the contribution to prediction uncertainty due to input data uncertainty, and since it also aligns well with the GUM framework~\cite{bipm2008,bipm2008supplement,bipm2011,balazs2008}.

The use of analytical expressions in this context offers a solution which is more accurate, transparent and reproducible than a Monte Carlo sampling approach. We have obtained analytical expressions for the mean and variance of the prediction output for a wide range of ML models, covering linear regression, penalised linear regression, support vector machines (SVMs), Gaussian Processes (GPs), kernel ridge regression and relevance vector machines (RVMs). 

For kernel-based models, our results are all for the case of RBF kernels. However, the restriction to RBF kernels is not too onerous since the RBF kernel is by far the most popular choice of kernel in general.

For linear models, only the mean and covariance of the input variables is required, and no assumptions about the distribution are needed. For kernel-based models, our methods require that input variable distributions can be modelled by certain statistical distributions such as multivariate Gaussians or independent uniform or symmetric triangular distributions. It is often sensible in metrology to model uncertainty using a statistical distribution. However, such a modelling assumption may be too restrictive for some applications, for example in~\cite{venton2021robustness} where the robustness of convolutional neural networks to physiological noise was investigated. In such cases, a sampling approach may be the only practical solution.

We also explored the computational tradeoffs between analytical and sampling-based approaches to uncertainty quantification. For linear models, we conclude that the analytical approach is always likely to be faster. For kernel-based models, the picture is more nuanced. We found that as the number of Monte Carlo trials is increased beyond roughly the number of training samples, it becomes the more computationally expensive method. This means that, for any given model, there exists an accuracy threshold, which increases as the size of the training set increases, such that it is more computationally expensive to obtain results with a Monte Carlo approach to an accuracy superior to this threshold.

Another factor to consider in the choice of method is the desired expression of prediction uncertainty. Except for the special case of linear models and multivariate Gaussian input variables, the analytical results presented here concern the mean and variance of the prediction output. Sometimes it is desired to characterise the output distribution in some other way, for example by computing quantiles and confidence/credible intervals. Distributional assumptions on the output distributions would need to be made to obtain such expressions of uncertainty based on the results in this paper. A Monte Carlo sampling approach, on the other hand, can also be used to compute quantiles, provided that care is taken to ensure a suitable number of samples is used.

In summary, in deciding on adopting an analytical or Monte Carlo sampling approach, it must be decided whether the the benefits of the analytical approach in terms of accuracy, transparency, reproducibility and (in some cases) computational efficiency outweigh the restrictions the analytical approach imposes on the modelling of input variables and the expression of output uncertainty.

\subsection{Future work}

We commented in Section~\ref{accuracy_validation} that an interesting direction for future investigation would be to analyse or improve the numerical stability of the analytical expressions given in this paper for kernel-based methods. The use of Taylor series expansions and floating-point algorithms~\cite{driscoll2014chebfun} may be useful in this respect.

We have considered in this paper some of the ML regression models that are most amenable to analytical uncertainty quantification. Two other popular classes of ML model are tree-based models (such as random forests) and neural networks. Obtaining analytical results for these models would be an interesting future challenge, though obtaining such results seems more challenging.

Finally, the results presented in Section~\ref{kernel_results} for kernel-based models and multivariate Gaussian input variables are a simplification of existing results in~\cite{candela2003propagation,quinonero2003prediction} which also take into account model uncertainty. This is achieved through a variance decomposition approach, and we note that such an approach could also be adopted with other models, and this is also left as future work.

\section*{Acknowledgements}

This work was supported by the UK Government’s Department for Science, Innovation and Technology (D-SIT). The Li-ion cell SOH and EIS data used in this paper was collected and released~\cite{chan2002release} as part of the 17IND10 \emph{LibForSecUse} project, which received funding from the EMPIR programme co-financed by the Participating States and from the European Union's Horizon 2020 research and innovation programme. The author wishes to thank Peter Harris (NPL), Edmund Dickinson (About:Energy) and Sam Bilson (NPL) for fruitful discussions. The author also wishes to thank Philip Aston (NPL and University of Surrey) for reviewing the paper and providing useful comments. 

\begin{appendix}

\section{Proofs of uncertainty propagation results for kernel-based models}\label{kernel_proofs}

\textbf{Proof of Theorem~\ref{general_kernel}}: 
The required expression for the mean of a point prediction of the form (\ref{kernel_point}) is given in ~\cite[Equation (30)]{quinonero2003prediction}. For the variance, we give a simplification of the argument in~\cite{quinonero2003prediction}. We have
$$\begin{array}{rcl}\textrm{Var} \,Y^{\ast}&=&\EE\left[\bal^{\top}\bk^{\ast}\right]^2-\left[\EE\,\bal^{\top}\bk^{\ast}\right]^2\\
&=&\bal^{\top}\EE\left[\bk^{\ast}\left(\bk^{\ast}\right)^{\top}\right]\bal-\left[\EE\,\bal^{\top}\bk^{\ast}\right]^2.
\end{array}
$$
The $i^{\textrm{th}}$ entry of $\bk^{\ast}$ is $k(\bx^{\ast},\bx_i)$ and the $ij^{\textrm{th}}$ entry of $\bk^{\ast}\left(\bk^{\ast}\right)^{\top}$ is $k(\bx^{\ast},\bx_i)k(\bx^{\ast},\bx_i)$. It then follows that the $i^{\textrm{th}}$ entry of $\EE\,\bk^{\ast}$ is given by
$$\int k(\bx,\bx_i)f(\bx)\textrm{d}\bx$$
and the $ij^{\textrm{th}}$ entry of $\EE\,\bk^{\ast}\left(\bk^{\ast}\right)^{\top}$ is given by
$$\int k(\bx,\bx_i) k(\bx,\bx_j)f(\bx)\textrm{d}\bx,$$
which gives the required result for the variance.\hfill$\Box$\\
\\
\textbf{Proof of Theorem~\ref{kernel_ind}}: From (\ref{l_def}) and (\ref{separable_RBF}), we have
$$\begin{array}{rcl}
l_i&=&\displaystyle\int k(\bx,\bx_i)f(\bx)\textrm{d}\bx\\
&=&\displaystyle\int \prod_{p=1}^m\exp\left[-\frac{(x^p-x_i^p)^2}{2\lambda_p^2}\right]\prod_{p=1}^m f^p(x^p)\textrm{d}\bx\\
&=&\displaystyle\prod_{p=1}^m\int \exp\left[-\frac{(x^p-x_i^p)^2}{2\lambda_p^2}\right]f^p(x^p)\textrm{d}x^p\\
&=&\displaystyle\prod_{p=1}^m l_i^p,\end{array}
$$
as required. From (\ref{L_def}) and (\ref{separable_RBF}), we have
$$\begin{array}{rcl}
L_{ij}&=&\displaystyle\int k(\bx,\bx_i)k(\bx,\bx_j)f(\bx)\textrm{d}\bx\\
&=&\displaystyle\int \prod_{p=1}^m\exp\left[-\frac{(x^p-x_i^p)^2}{2\lambda_p^2}\right]\prod_{p=1}^m\exp\left[-\frac{(x^p-x_j^p)^2}{2\lambda_p^2}\right]\prod_{p=1}^m f^p(x^p)\textrm{d}\bx\\
&=&\displaystyle\prod_{p=1}^m\int \exp\left[-\frac{(x^p-x_i^p)^2}{2\lambda_p^2}\right]\exp\left[-\frac{(x^p-x_j^p)^2}{2\lambda_p^2}\right]f^p(x^p)\textrm{d}x^p\\
&=&\displaystyle\prod_{p=1}^m L_{ij}^p,\end{array}
$$
which is also as required.\hfill$\Box$\\
\\
\textbf{Proof of Theorem~\ref{kernel_uniform}}: A uniform distribution on $[a,b]$ has pdf
\begin{equation}\label{uniform_pdf}
f(x)=\left\{\begin{array}{ll}
\displaystyle\frac{1}{b-a}&x\in[a,b]\\
0&\textrm{otherwise}.
\end{array}\right.
\end{equation}
It follows from (\ref{uniform_pdf}) that
$$\begin{array}{rcl}
l_i^p&=&\displaystyle\int_a^b\frac{1}{b-a} \exp\left[-\frac{(x^p-x_i^p)^2}{2\lambda_p^2}\right]\dx^p\\
&=&\displaystyle\frac{\lambda_p\sqrt{2\pi}}{b-a}\left[\Phi\left(\frac{x^p-x_i^p}{\lambda_p}\right)\right]_a^b.
\end{array}$$
(\ref{l_uniform}) now follows by evaluating the integral limits. On the other hand, it follows from (\ref{uniform_pdf}) that
$$L_{ij}^p=\displaystyle\int_a^b\frac{1}{b-a} \exp\left[-\frac{(x^p-x_i^p)^2}{2\lambda_p^2}\right]\exp\left[-\frac{(x^p-x_j^p)^2}{2\lambda_p^2}\right]\dx^p,$$
which can be rearranged to give
$$L_{ij}^p=\displaystyle\frac{1}{b-a}\int_a^b\exp\left[-\frac{(x^p-m_{ij}^p)^2}{\lambda_p^2}\right]\exp\left[-\frac{(x_i^p-x_j^p)^2}{4\lambda_p^2}\right]\dx^p.$$
The right-hand exponential is independent of $x^p$, and what is left is the integral of a univariate Gaussian pdf, and so we obtain
$$L_{ij}^p=\displaystyle\frac{1}{b-a}\exp\left[-\frac{(x_i^p-x_j^p)^2}{4\lambda_p^2}\right]\left[\lambda_p\sqrt{\pi}\Phi\left(\frac{x-m_{ij}^p}{(\lambda_p/\sqrt{2})}\right)\right]_a^b.$$
(\ref{L_uniform}) now follows by evaluating the integral limits.\hfill$\Box$\\
\\
\textbf{Proof of Theorem~\ref{kernel_tri}}: A symmetric triangular distribution on $[a,b]$ has pdf
\begin{equation}\label{triangular_pdf}
f(x)=\left\{\begin{array}{ll}
\displaystyle\frac{4(x-a)}{(b-a)^2}&a\le x\le \frac{a+b}{2}\\
\displaystyle\frac{4(b-x)}{(b-a)^2}&\frac{a+b}{2}\le x\le b\\
0&\textrm{otherwise}.
\end{array}\right.
\end{equation}
It follows from (\ref{separable_l_def}) and (\ref{triangular_pdf}) that
$$
l_i^p=\displaystyle\int_a^{\frac{a+b}{2}}\frac{4(x^p-a)}{(b-a)^2} \exp\left[-\frac{(x^p-x_i^p)^2}{2\lambda_p^2}\right]\dx^p+\int_{\frac{a+b}{2}}^b\frac{4(b-x^p)}{(b-a)^2} \exp\left[-\frac{(x^p-x_i^p)^2}{2\lambda_p^2}\right]\dx^p,$$
and the integrands can be split to give
$$\begin{array}{ll}l_i^p=&\displaystyle\int_a^{\frac{a+b}{2}}\frac{4(x^p-x_i^p)}{(b-a)^2} \exp\left[-\frac{(x^p-x_i^p)^2}{2\lambda_p^2}\right]\dx^p+\displaystyle\int_a^{\frac{a+b}{2}}\frac{4(x_i^p-a)}{(b-a)^2} \exp\left[-\frac{(x^p-x_i^p)^2}{2\lambda_p^2}\right]\dx^p\\
&+\displaystyle\int_{\frac{a+b}{2}}^b\frac{4(b-x_i^p)}{(b-a)^2} \exp\left[-\frac{(x^p-x_i^p)^2}{2\lambda_p^2}\right]\dx^p+\int_{\frac{a+b}{2}}^b\frac{4(x_i^p-x^p)}{(b-a)^2} \exp\left[-\frac{(x^p-x_i^p)^2}{2\lambda_p^2}\right]\dx^p,
\end{array}$$
and integrated to give
\begin{equation}\label{l_integral}
\begin{array}{ll}l_i^p=&\left[-\displaystyle\frac{4\lambda_p^2}{(b-a)^2}\exp\left\{-\frac{(x^p-x_i^p)^2}{2\lambda_p^2}\right\}\right]_a^{\frac{a+b}{2}}+\left[-\displaystyle\frac{4\lambda_p\sqrt{2\pi}(a-x_i^p)}{(b-a)^2}\Phi\left(\frac{x^p-x_i^p}{\lambda_p}\right)\right]_a^{\frac{a+b}{2}}\\
&+\left[\displaystyle\frac{4\lambda_p\sqrt{2\pi}(b-x_i^p)}{(b-a)^2}\Phi\left(\frac{x^p-x_i^p}{\lambda_p}\right)\right]_{\frac{a+b}{2}}^b+\left[\displaystyle\frac{4\lambda_p^2}{(b-a)^2}\exp\left\{-\frac{(x^p-x_i^p)^2}{2\lambda_p^2}\right\}\right]_{\frac{a+b}{2}}^b,
\end{array}
\end{equation}
from which (\ref{l_triangular}) now follows. On the other hand, it follows from (\ref{triangular_pdf}) that
\begin{equation}\label{L_integral1}
\begin{array}{ll}L_{ij}^p=&\displaystyle\int_a^{\frac{a+b}{2}}\frac{4(x^p-a)}{(b-a)^2} \exp\left[-\frac{(x^p-x_i^p)^2}{2\lambda_p^2}\right]\exp\left[-\frac{(x^p-x_j^p)^2}{2\lambda_p^2}\right]\dx^p\\
&+\displaystyle\int_{\frac{a+b}{2}}^b\frac{4(b-x^p)}{(b-a)^2} \exp\left[-\frac{(x^p-x_i^p)^2}{2\lambda_p^2}\right]\exp\left[-\frac{(x^p-x_j^p)^2}{2\lambda_p^2}\right]\dx^p.
\end{array}\end{equation}
By splitting and rearranging the integrals in (\ref{L_integral1}), we obtain
$$\begin{array}{ll}L_{ij}^p=&\displaystyle\int_a^{\frac{a+b}{2}}\frac{4(x^p-m_{ij}^p)}{(b-a)^2} \exp\left[-\frac{(x^p-m_{ij}^p)^2}{\lambda_p^2}\right]\exp\left[-\frac{(x_i^p-x_j^p)^2}{4\lambda_p^2}\right]\dx^p\\
&+\displaystyle\int_a^{\frac{a+b}{2}}\frac{4(m_{ij}^p-a)}{(b-a)^2} \exp\left[-\frac{(x^p-m_{ij}^p)^2}{\lambda_p^2}\right]\exp\left[-\frac{(x_i^p-x_j^p)^2}{4\lambda_p^2}\right]\dx^p\\
&+\displaystyle\int_{\frac{a+b}{2}}^b\frac{4(b-m_{ij}^p)}{(b-a)^2} \exp\left[-\frac{(x^p-m_{ij}^p)^2}{\lambda_p^2}\right]\exp\left[-\frac{(x_i^p-x_j^p)^2}{4\lambda_p^2}\right]\dx^p\\
&+\displaystyle\int_{\frac{a+b}{2}}^b\frac{4(m_{ij}^p-x^p)}{(b-a)^2} \exp\left[-\frac{(x^p-m_{ij}^p)^2}{\lambda_p^2}\right]\exp\left[-\frac{(x_i^p-x_j^p)^2}{4\lambda_p^2}\right]\dx^p.
\end{array}$$
and integration then gives
\begin{equation}\label{L_integral2}
\begin{array}{ll}L_{ij}^p=&
\left[-\displaystyle\frac{2\lambda_p^2}{(b-a)^2}\exp\left\{-\frac{(x^p-m_{ij}^p)^2}{\lambda_p^2}\right\}\exp\left\{-\frac{(x_i^p-x_j^p)^2}{4\lambda_p^2}\right\}\right]_a^{\frac{a+b}{2}}
\\
&+\left[-\displaystyle\frac{4\lambda_p\sqrt{\pi}(a-m_{ij}^p)}{(b-a)^2}\Phi\left(\frac{x^p-m_{ij}^p}{(\lambda_p/\sqrt{2})}\right)\exp\left\{-\frac{(x_i^p-x_j^p)^2}{4\lambda_p^2}\right\}\right]_a^{\frac{a+b}{2}}\\
&+\left[\displaystyle\frac{4\lambda_p\sqrt{\pi}(a-m_{ij}^p)}{(b-a)^2}\Phi\left(\frac{x^p-m_{ij}^p}{(\lambda_p/\sqrt{2})}\right)\exp\left\{-\frac{(x_i^p-x_j^p)^2}{4\lambda_p^2}\right\}\right]_{\frac{a+b}{2}}^b\\
&+\left[\displaystyle\frac{2\lambda_p^2}{(b-a)^2}\exp\left\{-\frac{(x^p-m_{ij}^p)^2}{\lambda_p^2}\right\}\exp\left\{-\frac{(x_i^p-x_j^p)^2}{4\lambda_p^2}\right\}\right]_{\frac{a+b}{2}}^b,
\end{array}
\end{equation}
from which (\ref{L_triangular}) now follows.\hfill$\Box$

\end{appendix}

\bibliography{refs}

\begin{thebibliography}{10}
\expandafter\ifx\csname url\endcsname\relax
  \def\url#1{\texttt{#1}}\fi
\expandafter\ifx\csname urlprefix\endcsname\relax\def\urlprefix{URL }\fi
\expandafter\ifx\csname href\endcsname\relax
  \def\href#1#2{#2} \def\path#1{#1}\fi

\bibitem{bilson2023machine}
S.~Bilson, A.~Thompson, D.~Tucker, J.~Pearce, A machine learning approach to
  automation and uncertainty evaluation for self-validating thermocouples, in:
  Proceedings of the International Temperature Symposium (ITS-10), San Diego,
  USA, 2023.

\bibitem{chan2022comparison}
H.~S. Chan, E.~Dickinson, T.~Heins, J.~Park, M.~Gaber{\v{s}}{\v{c}}ek, Y.~Lee,
  M.~Heinrich, V.~Ruiz, E.~Napolitano, P.~Kauranen, et~al., Comparison of
  methodologies to estimate state-of-health of commercial {L}i-ion cells from
  electrochemical frequency response data, Journal of Power Sources 542 (2022)
  231814.

\bibitem{coquelin2019towards}
L.~Coquelin, N.~Fischer, N.~Feltin, L.~Devoille, G.~Felhi, Towards the use of
  deep generative models for the characterization in size of aggregated
  $\mathrm{TiO_2}$ nanoparticles measured by {S}canning {E}lectron {M}icroscopy
  ({SEM}), Materials Research Express 6~(8) (2019).

\bibitem{lary2018machine}
D.~Lary, G.~Zewdie, X.~Liu, D.~Wu, E.~Levetin, R.~Allee, N.~Malakar, A.~Walker,
  H.~Mussa, A.~Mannino, et~al., Machine learning applications for earth
  observation, Earth observation open science and innovation 165 (2018).

\bibitem{robinson2023impact}
S.~Robinson, P.~Harris, S.-H. Cheong, L.~Wang, V.~Livina, G.~Haralabus,
  M.~Zampolli, P.~Nielsen, Impact of the {COVID}-19 pandemic on levels of
  deep-ocean acoustic noise, Scientific Reports 13~(1) (2023) 4631.

\bibitem{mathmet2024SRA}
E.~M.~N. for Mathematics, Statistics, Strategic research agenda 2023 – 2033,
  \url{https://www.euramet.org/european-metrology-networks/mathmet/strategy/strategic-research-agenda},
  accessed: 25-04-2024 (2024).

\bibitem{balazs2008}
BIPM, IEC, IFCC, ILAC, ISO, IUPAC, IUPAP, OIML, International vocabulary of
  metrology -- {B}asic and general concepts and associated terms ({VIM}), Joint
  Committee for Guides in Metrology (JCGM)3rd edition (2008 version with minor
  corrections).

\bibitem{bipm2008}
BIPM, IEC, IFCC, ILAC, ISO, IUPAC, IUPAP, OIML, Evaluation of measurement data
  -- {G}uide to the expression of uncertainty in measurement ({GUM} 1995 with
  minor corrections), Joint Committee for Guides in Metrology (JCGM) 100
  (2008).

\bibitem{bipm2008supplement}
BIPM, IEC, IFCC, ILAC, ISO, IUPAC, IUPAP, OIML, Evaluation of measurement data
  -- {S}upplement 1 to the `{G}uide to the expression of uncertainty in
  measurement', Joint Committee for Guides in Metrology (JCGM) 101 (2008).

\bibitem{bipm2011}
BIPM, IEC, IFCC, ILAC, ISO, IUPAC, IUPAP, OIML, Evaluation of measurement data
  -- {S}upplement 2 to the `{G}uide to the expression of uncertainty in
  measurement' -- {E}xtension to any number of output quantities, Joint
  Committee for Guides in Metrology (JCGM) 102 (2011).

\bibitem{venton2021robustness}
J.~Venton, P.~Harris, A.~Sundar, N.~Smith, P.~Aston, Robustness of
  convolutional neural networks to physiological electrocardiogram noise,
  Philosophical Transactions of the Royal Society A 379~(2212) (2021) 20200262.

\bibitem{forbes2002generalised}
A.~Forbes, P.~Harris, I.~Smith, Generalised gauss-markov regression, in:
  Algorithms for Approximation IV, University of Huddersfield, 2002, pp.
  270--277.

\bibitem{klauenberg2022gum}
K.~Klauenberg, S.~Martens, A.~Bo{\v{s}}njakovi{\'c}, M.~Cox, A.~van~der Veen,
  C.~Elster, The gum perspective on straight-line errors-in-variables
  regression, Measurement 187 (2022) 110340.

\bibitem{martin2023aleatoric}
J.~Martin, C.~Elster, Aleatoric uncertainty for errors-in-variables models in
  deep regression, Neural Processing Letters 55~(4) (2023) 4799--4818.

\bibitem{candela2003propagation}
J.~Quinonero-Candela, A.~Girard, J.~Larsen, C.~Rasmussen, Propagation of
  uncertainty in {B}ayesian kernel models-application to multiple-step ahead
  forecasting, in: 2003 IEEE International Conference on Acoustics, Speech, and
  Signal Processing, 2003. Proceedings.(ICASSP'03)., Vol.~2, 2003, pp. II--701.

\bibitem{quinonero2003prediction}
J.~Quinonero-Candela, A.~Girard, C.~Rasmussen, Prediction at an uncertain input
  for {G}aussian {P}rocesses and relevance vector machines-application to
  multiple-step ahead time-series forecasting, Tech. rep., Technical University
  of Denmark, DTU: Informatics and Mathematical Modelling (2003).

\bibitem{wan2014analytical}
H.~Wan, Z.~Mao, M.~Todd, W.~Ren, Analytical uncertainty quantification for
  modal frequencies with structural parameter uncertainty using a gaussian
  process metamodel, Engineering Structures 75 (2014) 577--589.

\bibitem{hastie2009elements}
T.~Hastie, R.~Tibshirani, J.~Friedman, The elements of statistical learning:
  data mining, inference, and prediction, Springer Science \& Business Media,
  2009.

\bibitem{tibshirani1996regression}
R.~Tibshirani, Regression shrinkage and selection via the {LASSO}, Journal of
  the Royal Statistical Society: Series B (Methodological) 58~(1) (1996)
  267--288.

\bibitem{johnson2002applied}
R.~Johnson, D.~Wichern, et~al., Applied multivariate statistical analysis,
  Prentice Hall Upper Saddle River, NJ, 2002.

\bibitem{williams2006gaussian}
C.~Williams, C.~Rasmussen, Gaussian {P}rocesses for machine learning, MIT press
  Cambridge, MA, 2006.

\bibitem{welling2013kernel}
M.~Welling, Max {W}elling’s classnotes in machine learning - kernel ridge
  regression,
  \url{https://web2.qatar.cmu.edu/\~gdicaro/10315-Fall19/additional/welling-notes-on-kernel-ridge.pdf},
  accessed: 19-12-2023 (2013).

\bibitem{tipping2001sparse}
M.~Tipping, Sparse {B}ayesian learning and the relevance vector machine,
  Journal of machine learning research 1~(Jun) (2001) 211--244.

\bibitem{kotak2021end}
Y.~Kotak, C.~Marchante~Fern{\'a}ndez, L.~Canals~Casals, B.~S. Kotak, D.~Koch,
  C.~Geisbauer, L.~Trilla, A.~G{\'o}mez-N{\'u}{\~n}ez, H.-G. Schweiger, End of
  electric vehicle batteries: {R}euse vs. recycle, Energies 14~(8) (2021) 2217.

\bibitem{shahjalal2022review}
M.~Shahjalal, P.~Roy, T.~Shams, A.~Fly, J.~Chowdhury, M.~Ahmed, K.~Liu, A
  review on second-life of {L}i-ion batteries: prospects, challenges, and
  issues, Energy 241 (2022) 122881.

\bibitem{chan2002release}
H.~Chan, E.~Dickinson, E.~Fedorovskaya, M.~Gaber{\v{s}}{\v{c}}ek, T.~Heins,
  N.~Meddings, Y.~Lee, J.~Mo{\v{s}}kon, J.~Park, V.~Ruiz, S.~Seitz,
  Libforsecuse data release - impedance spectra of life cycle tests of
  commercial 18650 cells, \url{https://zenodo.org/records/6418665} (2002).

\bibitem{walck1996hand}
C.~Walck, et~al., Hand-book on statistical distributions for experimentalists,
  Stockholms Universitet, 1996.

\bibitem{bachstein2019uncertainty}
S.~Bachstein, Uncertainty quantification in deep learning, Master's thesis, Ulm
  University (2019).

\bibitem{thompson2021uncertainty}
A.~Thompson, K.~Jagan, A.~Sundar, R.~Khatry, et~al., Uncertainty evaluation for
  machine learning, Tech. rep., National Physical Laboratory, technical report
  MS-34 (2021).

\bibitem{wubbeler2010two}
G.~W{\"u}bbeler, P.~Harris, M.~Cox, C.~Elster, A two-stage procedure for
  determining the number of trials in the application of a {M}onte {C}arlo
  method for uncertainty evaluation, Metrologia 47~(3) (2010) 317.

\bibitem{driscoll2014chebfun}
T.~Driscoll, N.~Hale, L.~Trefethen, Chebfun guide 1st edition,
  \url{https://www.chebfun.org/docs/guide/chebfun\_guide.pdf}, accessed:
  19-12-2023 (2014).

\end{thebibliography}
\bibliographystyle{elsarticle-num}

\end{document}